\documentclass{article}

\usepackage[preprint]{neurips_2025}

\usepackage[utf8]{inputenc} 
\usepackage[T1]{fontenc}    
\usepackage{hyperref}       
\usepackage{url}            
\usepackage{booktabs}       
\usepackage{amsfonts}       
\usepackage{nicefrac}       
\usepackage{microtype}      
\usepackage{xcolor}         
\usepackage{graphicx}
\usepackage{amsmath}
\usepackage{longtable}
\usepackage{wrapfig}
\usepackage{caption}  
\usepackage{soul}
\usepackage{comment}

\newcommand{\dataset}[1]{\textsc{#1}}

\title{Training Transformers for Mesh-Based Simulations}

\author{
  \AND
  Paul Garnier\thanks{Corresponding author}\\
  Mines Paris - PSL University\\
  Centre for Material Forming (CEMEF) \\
  CNRS \\
  \texttt{paul.garnier@minesparis.psl.eu}\\
  \And
  Vincent Lannelongue\\
  Mines Paris - PSL University\\
  Centre for Material Forming (CEMEF) \\
  CNRS \\
  \texttt{vincent.lannelongue@minesparis.psl.eu}\\ 
  \And
  Jonathan Viquerat\\
  Mines Paris - PSL University\\
  Centre for Material Forming (CEMEF) \\
  CNRS \\
  \texttt{jonathan.viquerat@minesparis.psl.eu}\\  
  \And
  Elie Hachem\\
  Mines Paris - PSL University\\
  Centre for Material Forming (CEMEF) \\
  CNRS \\
  \texttt{elie.hachem@minesparis.psl.eu}\\
}

\begin{document}

\maketitle

\begin{abstract}
Simulating physics using Graph Neural Networks (GNNs) is predominantly driven by message-passing architectures, which face challenges in scaling and efficiency, particularly in handling large, complex meshes. These architectures have inspired numerous enhancements, including multigrid approaches and $K$-hop aggregation (using neighbours of distance $K$), yet they often introduce significant complexity and suffer from limited in-depth investigations. In response to these challenges, we propose a novel Graph Transformer architecture that leverages the adjacency matrix as an attention mask. The proposed approach incorporates innovative augmentations, including Dilated Sliding Windows and Global Attention, to extend receptive fields without sacrificing computational efficiency. Through extensive experimentation, we evaluate model size, adjacency matrix augmentations, positional encoding and $K$-hop configurations using challenging 3D computational fluid dynamics (CFD) datasets. We also train over 60 models to find a scaling law between training FLOPs and parameters. The introduced models demonstrate remarkable scalability, performing on meshes with up to 300k nodes and 3 million edges. Notably, the smallest model achieves parity with MeshGraphNet while being $7\times$ faster and $6\times$ smaller. The largest model surpasses the previous state-of-the-art by $38.8$\% on average and outperforms MeshGraphNet by $52$\% on the all-rollout RMSE, while having a similar training speed. Code and datasets are available at \url{https://github.com/DonsetPG/graph-physics}.
\end{abstract}

\section{Introduction}
\label{sec:introduction}

Simulating physical phenomena, particularly in computational fluid dynamics (CFD), involves solving partial differential equations over complex domains represented as unstructured meshes \cite{HACHEM20108643}. These simulations typically require intensive linear algebra computations distributed across multiple processors. Moreover, each new simulation is performed independently, disregarding insights gained from previous runs, which motivates the integration of machine learning (ML) techniques for physics simulation.

Early ML models for physics focused on structured grids, using convolutional neural networks (CNNs) \cite{Tompson2016,Thuerey2018,Chen2019} or generative adversarial networks (GANs) \cite{Chu_2017} to predict velocity fields and other physical quantities. Several enhancements were proposed to integrate physical constraint by  adding residual equations into the model's loss function, forming Physically-Informed Neural Networks (PINN) \cite{RAISSI2019686}. While effective, these image-based methods are inherently limited to structured grids, preventing local refinement and complicating applications to 3D scenarios. 

\begin{figure*}[tbh!]
  \centering
  \includegraphics[width=1\textwidth]{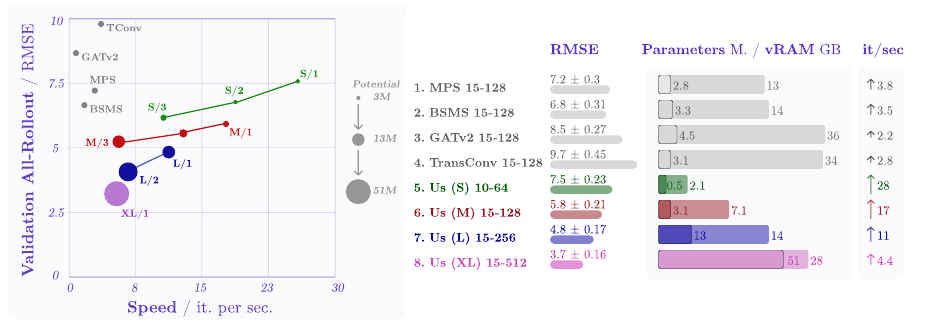}
  \caption{\small\textbf{RMSE over autoregressive trajectories on the \dataset{Coarse Aneurysm} Dataset.} We showcase a Message Passing approach (MPS with 15 layers of width 128), a bi-stride (BSMS), 2 attention based models and our architecture. \textbf{(left)} Performances are shown without any masking pre-training and with a standard Adjacency Matrix without augmentations. Our new architecture constantly beats the previous SOTA based on message-passing architecture. \textbf{(right)} Comparison between other architectures and our approach in terms of validation RMSE, number of parameters, vRAM consumption and training speed. Even our largest model is faster than a much smaller MPS model.}
  \label{fig:overview}
\end{figure*}

To address this, graph neural networks (GNNs) emerged as a natural fit for unstructured meshes, with message-passing architectures (MPS) \cite{Battaglia2018}, \cite{sanchezgonzalez2020learning,pfaff2021learning, Libao2023} enabled simulations directly on graph-based representations. MPS-based GNNs have advanced physics simulation in diverse domains, from weather forecasting \cite{lam2023graphcastlearningskillfulmediumrange} to blood flow modeling \cite{Suk_2024}. To face scaling limitations and significant computational costs, the MPS architecture has seen many improvements, such as Multigrid (an approach based on FEM \cite{John2002} that uses a hierarchy of mesh discretizations in order to propagate information at different scales) \cite{lin2024upsamplingonlyadaptivemeshbasedgnn, Fortunato2022, lino2021simulating, Yang2022amgnet, taghibakhshi2023mggnn, cao2023efficientlearningmeshbasedphysical}. Most methods applying graph neural networks to simulate physics have now significantly gained in complexity, with different variants of MPS at different mesh scales.

Simultaneously, the Transformer architecture \cite{vaswani2023attention} has revolutionized natural language processing (NLP)  \cite{devlin2019bert, gpt2} and computer vision with the creation of Vision Transformers (ViT) \cite{dosovitskiy2021image, arnab2021vivit}, introducing attention mechanisms that scale efficiently and effectively capture long-range dependencies. 

Many improvements over those methods have been developed, such as a Masking pre-training technique for ViTs, similar to the Cloze task \cite{taylor1953cloze} for NLP \cite{he2021masked}, or on the attention masking with the introduction of Sliding Windows \cite{beltagy2020longformerlongdocumenttransformer, zaheer2021bigbirdtransformerslonger,jiang2023mistral7b}.

Recent adaptations of Transformers for GNNs have been explored \cite{yun2020graph, müller2023attending}, with the addition of attention directly inside message passing \cite{veličković2018graph, shi2021masked}. Another approach was to reduce computation complexity by using the graph topology instead of attending all nodes to all nodes \cite{dwivedi2021generalizationtransformernetworksgraphs, ying2021transformersreallyperformbad}. Similar questions to NLP regarding Positional Encoding (PE) also emerged, with work leaning on the usage of laplacian eigen vectors, learnable vectors or random walks \cite{dwivedi2022graphneuralnetworkslearnable,dwivedi2022benchmarkinggraphneuralnetworks,kreuzer2021rethinkinggraphtransformersspectral}. More recent approaches use attention to aggregate nodes \cite{janny2023eaglelargescalelearningturbulent} or to support Neural Operator frameworks \cite{alkin2025universalphysicstransformersframework}. 

Similar to the improvements made for Transformers and ViTs, methods have also been developed to reduce computation while increasing receptive fields, such as Mix-Hops \cite{abuelhaija2019mixhophigherordergraphconvolutional} and edge jumping \cite{gladstone2023gnnbasedphysicssolvertimeindependent, xu2018representationlearninggraphsjumping}. Pre-training methods similar to the Cloze tasks were developed as well by masking either subset of nodes or edges \cite{Tan2022,Hu2020gptgnn,garnier2024meshmask,zhou2024masked}.

In this work, we introduce a novel Transformer-based GNN architecture that directly utilizes the adjacency matrix as a sparse attention mask. This straightforward approach eliminates the need for traditional message-passing while preserving the graph structure. This leads to a very different formulation than GAT \cite{veličković2018graph} or \cite{shi2021maskedlabelpredictionunified} where edge features are integrated and a dense representation of the $QK^T$ matrix is used. The closest formulation to our approach can be found in \cite{dwivedi2021generalizationtransformernetworksgraphs} where the Hadamard product is applied inside instead of outside the softmax function and \cite{wu2024transolverfasttransformersolver} where physics-aware tokens and projection on different slices are used.

We motivate the usage of transformers for graph tasks by their strong performances in other fields, the similarities in terms of data structures, and more importantly, by the belief that transformers are actually a particular case of GNNs \cite{barbero2024transformersneedglassesinformation}. By augmenting the adjacency matrix with Dilated Sliding Windows, Global Attention, and Random Connections, we also introduces a novel method for expanding the receptive field, distinct from traditional multigrid or mesh coarsening techniques. 

While it is true that our Dilation Augmentation can be related to $K$-hop message passing, our approach offers different sizes of receptive fields for different attention heads, making it more versatile than simply increasing the hop size of a layer. This is also the first time our other two augmentations are being adapted to Graph Neural Networks. 

We performed a comprehensive ablation study to evaluate key design choices, including the model size, the number of neighboring nodes considered in the attention mechanism, the type of positional encoding, and the augmentation strategies for the adjacency matrix. This analysis was conducted on a dataset of greater complexity than the commonly used flow past a cylinder, using larger 3D meshes with over 10,000 nodes. Furthermore, our models were trained on even larger meshes, scaling up to 300,000 nodes and 3 million edges. These experiments demonstrate the scalability and robustness of the approach, even when applied to very large mesh.

Moreover, we propose an extensive experiments consisting of more than 60 models to reveal scaling laws between model size and training FLOPs, similar to \cite{kaplan2020scalinglawsneurallanguage, hoffmann2022trainingcomputeoptimallargelanguage}, offering practical guidance for deploying the proposed architecture across diverse datasets and mesh sizes. 

Despite its simplicity, the proposed architecture (\textbf{XL/1})\footnote{51m parameters} outperforms the current State-of-the-art (SOTA) models on various challenging physics datasets by $38.8$\% on average. The smallest model (\textbf{S/1})\footnote{500k parameters} matches the performance of MeshGraphNet \cite{pfaff2021learning} while being $7\times$ faster and $6\times$ smaller (see \autoref{fig:overview}). Additionally, we highlight the performance of the transformers on non-physics datasets in \autoref{sec:noncfd} for completeness. Code and datasets will be released upon publication.

\section{Model Architecture}
\label{sec:architecture}

\subsection{Mesh as Graph}
\label{subsec:graph}

We consider a mesh as an undirected graph $\mathcal{G} = (\mathcal{V},\mathcal{E})$. 
$\mathcal{V} = \{\mathbf{x}_i\}_{i=1:N}$ is the set of nodes, where each $\mathbf{x}_i \in \mathbb{R}^{p}$ represents the attributes of node $i$.
$\mathcal{E} = \{\left(\mathbf{e}_k, r_k, s_k\right)\}_{k=1:N^e}$ is the set of edges, where each $\mathbf{e}_k$ represents the attributes of edge $k$, $r_k$ is the index of the receiver node, and $s_k$ is the index of the sender node. 

In the remaining of our architecture, we omit the attributes of the edges and consider each node as a token. We note $\mathbf{X} = (\mathbf{x}_1, \mathbf{x}_2,...,\mathbf{x}_N)^{T} \in \mathbb{R}^{N \times p}$ our input matrix, made of $N$ tokens of dimension $p$. Let $\mathbf{Z} = (\mathbf{z}_1, \mathbf{z}_2,...,\mathbf{z}_N)^{T} \in \mathbb{R}^{N \times d}$ be the $d$-dimensional embedding of our nodes. We define $\mathbf{A}$ as the adjacency matrix of our graph, setting $\mathbf{A}_{ij} = \mathbf{A}_{ji} = 1$ if $(i,j) \in \mathcal{E}$ and $0$ otherwise.

\subsection{Model}
\label{subsec:model}

Our model follows an Encode-Process-Decode architecture similar to \cite{Battaglia2018}. The Encoder maps the input nodes into a latent space. We then apply $L$ layers of our transformer architecture. Finally, the Decoder maps back our outputs into a meaningful space. At each step, our model is auto-regressive, meaning that the output of our model is used as an input for the next step of the simulation.

\paragraph{Encoder and Decoder}
\label{para:encoder-decoder}

We simplify the architecture from \cite{pfaff2021learning} by using only two linear layers to encode our inputs. We also use only a node encoder since our model does not use edge attributes. Our encoder maps our nodes $\mathbf{X} \in \mathbb{R}^{N \times p}$ into a latent space $\mathbf{Z} \in \mathbb{R}^{N \times d}$. The parameter $d$ is shared across all our layers as the main width parameter. The Decoder generates an output from the latent space using two linear layers.

\subsubsection{Processor}
\label{subsubsec:processor}

Our Processor is made of $L$ Transformer Blocks. Each block takes the latest latent representation $\mathbf{Z}$ and the Adjacency matrix $\mathbf{A}$ as input. The Adjacency matrix is usually augmented following strategies detailed in \autoref{sec:adjacency}. Each block has two sub-layers: a Masked Multi-Head Self-Attention layer and a Gated MLP layer. We also add residual connections around those two layers, following the original Transformer implementation from \cite{vaswani2023attention}. We follow those connections with Layer Normalization using RMSNorm \cite{zhang2019rootmeansquarelayer}. An overview of the architecture is presented in \autoref{fig:archi}.

\paragraph{Masked Multi-Head Self-Attention}
\label{para:mmha}

We use the original implementation from \cite{vaswani2023attention}, with the Adjacency Matrix as a mask when computing $QK^T$ using the Hadamard product. Leaving the head dimension out for clarity, we perform the following operation in each layer: 

\begin{equation}
    \text{Attention}(\mathbf{Z}) = \left(\mathbf{A}\odot\text{softmax}\Big( \frac{QK^T}{\sqrt{d}} \Big)\right)V
\end{equation}

where $Q, K, V$ are linear projections of $\mathbf{Z}$. Finally, we project the results to $\mathbb{R}^{N \times d}$ by multiplying $\text{Attention}(\mathbf{Z})$ by $W_o \in \mathbb{R}^{d \times d}$.

\paragraph{Gated MLP}
\label{para:gated}

We then pass $\mathbf{Z}$ into a Gated Multi-Layer Perceptron \cite{dauphin2017language} with GeLU non-linearity \cite{hendrycks2023gaussian} which is now standard in modern transformers \cite{shazeer2020gluvariantsimprovetransformer,de2024griffinmixinggatedlinear}. The GatedMLP processes $\mathbf{Z}$ with two parallel linear projections of size $e \times d$ (we use $e=3$ in all experiments), followed by a GeLU function on one of the branches. Finally, the two branches are merged with an Hadamard product before being passed in a final linear layer of size $d$:

\begin{equation}
    \mathbf{Z} = W_f\Big(\text{GeLU}\big(W_l \mathbf{Z} + b_l\big) \odot (W_r \mathbf{Z} + b_r)\Big) + b_f
\end{equation}

We can summarize our architecture as follows:

\begin{align}
    \mathbf{Z}_0 &= \text{MLP}(\mathbf{X}) && \mathbf{X} \in \mathbb{R}^{N \times p}, \mathbf{Z} \in \mathbb{R}^{N \times d} \label{eq:embedding} \\
    \mathbf{Z'}_l &= \text{RMSNorm}\big(\text{MMHA}(\mathbf{Z}_{l-1}, \mathbf{A}) + \mathbf{Z}_{l-1}\big) && \ell \in [1\ldots L] \label{eq:msa_apply} \\
    \mathbf{Z}_l &= \text{RMSNorm}\big(\text{GatedMLP}(\mathbf{Z'}_{l}) + \mathbf{Z'}_{l}\big) && \ell \in [1\ldots L] \label{eq:mlp_apply} \\
    \mathbf{y} &= \text{MLP}(\mathbf{Z}_L) \label{eq:final_rep}
\end{align}

\subsection{Positional Encoding}
\label{subsec:pe}

If positional encoding is a significant question in the case of NLP, it is also the case for graphs that do not hold any geometric information in their node's features. In our datasets, each node holds its 2D or 3D coordinates, and we study whether those attributes are essential in the ablation study. We also studied variations of this positional encoding with Laplacien Eigen Vectors \cite{dwivedi2021generalizationtransformernetworksgraphs}, learnable positional Encoding \cite{kreuzer2021rethinkinggraphtransformersspectral} and RandomWalk \cite{dwivedi2022graphneuralnetworkslearnable}. For our CFD datasets, we found no improvements in comparison to the simple geometric coordinates. We hypothesize that the explicit geometric information provided by the node coordinates is more directly relevant for modeling physical systems. In the case of non-CFD datasets \autoref{sec:noncfd}, we show that our architecture is robust even without any Positional Encoding.

\section{Augmentations of the Adjacency Matrix}
\label{sec:adjacency}

\begin{figure}
  \centering
  \includegraphics[width=\columnwidth]{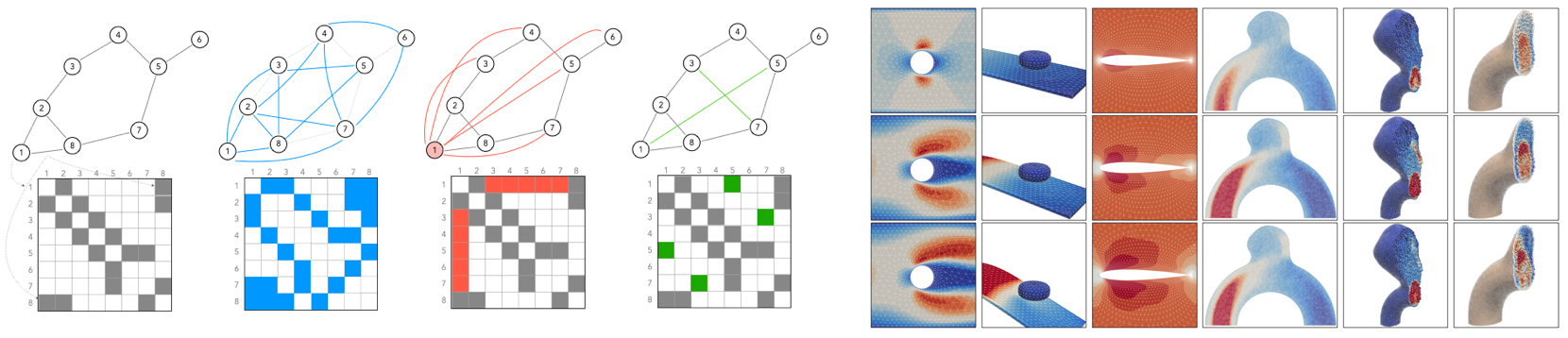}
  \caption{\small\textbf{(left) Details of the Augmented Adjacency matrices} \textbf{(1)} Default Adjacency Matrix $A$. \textbf{(2)} Adjacency Matrix with a Dilation of size $2$. \textbf{(3)} Adjacency Matrix with Global Attention based on node $1$. \textbf{(4)} Adjacency matrix with Random Jumpers between nodes $(1,5)$ and $(3,7)$.
  \textbf{(right) Dataset Overview} From left to right: \dataset{Cylinder}, \dataset{Plate}, \dataset{Airfoil}, \dataset{2D-Aneurysm}, \dataset{3D-CoarseAneurysm} and \dataset{3D-Aneurysm}.}
  \label{fig:adjacency}
\end{figure}

The original transformer has a complexity of $O(n^2)$ (with $n$ being the sequence length, or the number of nodes in our case) with every node attending to every node. We move from that paradigm by using a natural formulation when working with graphs and use the Adjacency Matrix as a mask. This is similar to using a Sliding Window of width $w$ with $w$ the maximum degree $\textit{deg}_{max}$ of our graph. This leads to a complexity of $O(n\times \textit{deg}_{max})$. We improve this adjacency matrix by drawing inspiration from \cite{beltagy2020longformerlongdocumenttransformer, zaheer2021bigbirdtransformerslonger} and introduce an Augmented Version of the Adjacency Matrix incorporating Random Connections, Dilation and Global Attention. A schematic representation of these augmentations is provided in \autoref{fig:adjacency}.

We find that these methods enable information to propagate much further compared to the regular Adjacency Matrix. Moreover, they are task-agnostic and independent of the specific graph structure. At the finite element level, they do not require complex mesh manipulations such as dynamic multigrid or mesh coarsening beforehand. We provide a theoretical analysis of using attention and an adjacency matrix instead of a regular message passing architecture in \autoref{annex:th-consideration}.

\textbf{Dilated Adjacency} To increase the number of neighbors seen by each node, one can increase the size of the $K$-hop. For example with $K=2$, one simply has to perform the same operations but with $\mathbf{A}' = \mathbf{A} + \mathbf{A}^2$. This leads to a complexity of $O(n\times \textit{deg}_{max}^K)$. In order to increase the receptive field in a similar fashion without increasing the number of edges too much, one can define a Dilated Adjacency Matrix. Instead of using $\mathbf{A}$, we chose to use $\mathbf{A}^k$. Similar to \cite{beltagy2020longformerlongdocumenttransformer}, we let our model focus on different sizes of receptive fields by setting different adjacency matrices for different heads. This is similar to the usual definition of $K$-hop message passing from \cite{feikhop} where each hop gets assigned a different processing function. We tried three configurations on a model with $L=15$ layers: \emph{2-Dilation} which uses $\mathbf{A}^2$ on half of the heads in the last 5 layers,  \emph{3-Dilation} which uses $\mathbf{A}^3$ on half of the heads in the last 5 layers and \emph{2-3-Dilation} which uses $\mathbf{A}^2$ on half of the heads in layers 5-10 and $\mathbf{A}^3$ on half of the heads in the last 5 layers. 

\textbf{Adjacency with Random Connections} To address the inherent locality of updates in our architecture, we introduce random edges to enhance the geometrical length of information flow. For a given number $j$, we randomly select $j$ pairs of nodes and add an edge between them. This approach enables information to propagate much further at a very low cost. Notably, these edges are generated randomly at each training and inference step, meaning they are neither pre-computed nor stored. Our experiments reveal that the model can effectively learn to utilize these new edges regardless of their locations. Furthermore, we observe that with a sufficient number of added edges, information flow is significantly improved, even when many of the edges are not placed in meaningful locations. This approach results in a complexity of $O(n^{1+\lambda})$, where $\lambda \in [0,1]$ denotes the proportion of random nodes considered.

\textbf{Adjacency with Global Attention} Similar to how certain words hold more significance than others in a task, some nodes are more important than others within a graph. To address this, we introduce Global Attention by symmetrically connecting specific nodes to every other node in the Adjacency Matrix. For instance, in the \dataset{Cylinder} dataset, the nodes forming the cylinder are deemed the most important, while in the \dataset{Aneurysm} dataset, the inflow boundary nodes are prioritized. To mitigate the potential for excessive connections, we select only a random sample\footnote{during our ablation study, we choose $1\%$, $5\%$ or $10\%$} of those global nodes. \autoref{appendix:datasets} provides insight into the choice of the considered nodes for each dataset. This approach results in a complexity of $O(n^{1+\lambda})$, where $\lambda \in [0,1]$ denotes the proportion of global nodes considered.

\section{Datasets}
\label{sec:datasets}

We evaluated our models on different use cases. Details such as the attributes used and the simulation time step $\Delta t$ can be seen in Table \ref{tab:datasets-details} and \autoref{fig:adjacency}. Each training set contains 100 trajectories, and the testing set contains 20 trajectories. Datasets from the COMSOL solver are initially from \cite{pfaff2021learning}. The \dataset{3D Aneurysm} dataset is from \cite{aneurysmdataset} and built using \cite{cimlib}. \dataset{3D Aneurysm} represents a large gap in comparison with the other datasets (more than 200k nodes on average) and makes for a very challenging task. From this dataset, we generated two others, one in 2D made from slices (\dataset{2D-Aneurysm}) and one with a coarser mesh (\dataset{3D-CoarseAneurysm}). Every model is trained with an $L_2$ loss average on all nodes. To evaluate our models, we use the 1-step RMSE and the All-Rollout RMSE defined in \cite{pfaff2021learning} and in \autoref{appendix:metrics}.

\section{Training}
\label{sec:training}

Our ablation study focuses on four base models, detailed in \autoref{tab:all_models}.  Models were trained for varying epochs (10 to 30), varying adjacency matrices, and different positional encoding. We focus our ablation study not on the usual \dataset{Cylinder} dataset but on the \dataset{3D-CoarseAneurysm} dataset. After running numerous experiments on both datasets, we found the \dataset{Cylinder} dataset too simple for our models, preventing sufficient gaps from appearing. We conducted a tuning of the learning rate value and schedule (exponential decay or warmup and cosine decay) on our smallest model (\textbf{S/1}) and interpolated from it for our bigger models, similar to \cite{wortsman2023smallscaleproxieslargescaletransformer}. We detail those experiments in \autoref{subappendix:other-differences}. Overall, we find that models trained with warmup and cosine decay perform better but that our bigger model needs a longer warmup (up to 5k iterations for \textbf{L} and \textbf{XL} models). Similarly, while all our models see stable training with a maximal learning rate of $10^{-3}$, our \textbf{XL} model needs a maximal learning rate of $10^{-4}$.

We also compare in \autoref{subappendix:other-differences} differences between an Adam \cite{kingma2017adam} and an AdamW \cite{loshchilov2019decoupledweightdecayregularization} optimizer. We find the AdamW optimizer to perform better and use it in all our training, with parameters $\beta _1 = 0.9, \beta _2 = 0.95$, and $\text{weight\_decay} = 10^{-4}$. We give details in \autoref{subsec:scaling} about the optimal model size and number of training iterations by finding a scaling law similar to \cite{kaplan2020scalinglawsneurallanguage} and use them to train all of our final models. We introduce noise to our inputs using the same strategy as \cite{sanchezgonzalez2020learning}.
More specifically, we add random noise $\mathcal{N}(0,\sigma)$ to dynamical variables (see \autoref{tab:noises}) at each training step and train solely on next-step prediction.

During our ablation study, we do not use any pre-training techniques. However, to train our final models, we stack 2 of our transformers models in an Encoder-Decoder fashion and pretrain them on Masked version of graphs, following the work from \cite{garnier2024meshmask}. After this pre-training phase, we discard the Decoder and finetune our Encoder (a transformer model) on unmasked graphs.

\begin{figure*}[tbh!]
  \centering
  \includegraphics[width=0.99\textwidth]{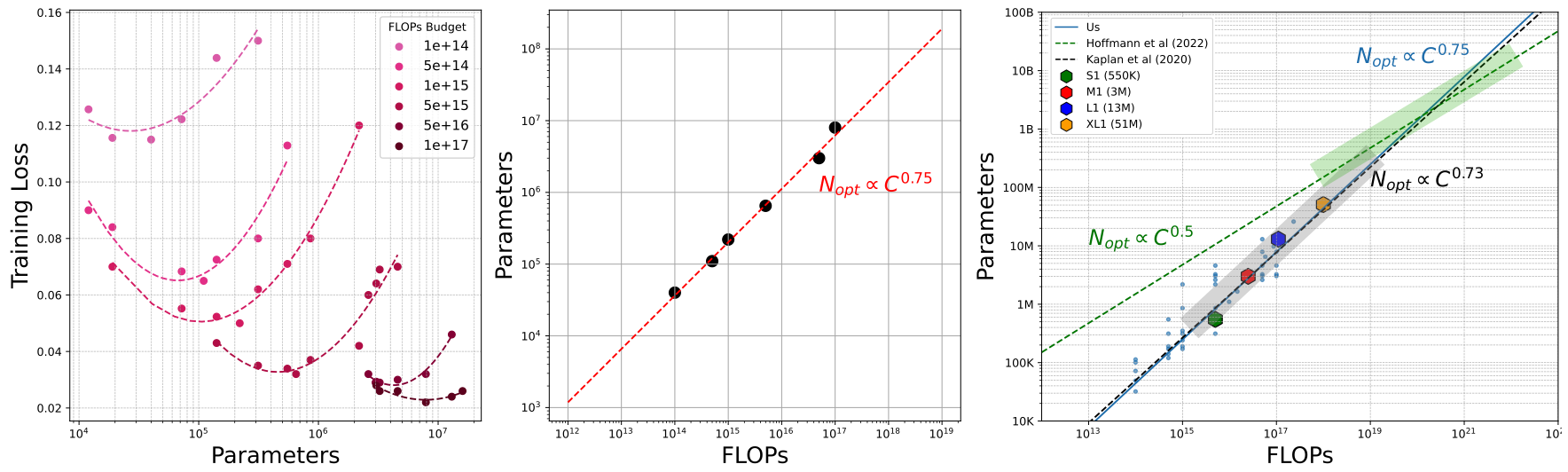}
  \caption{\small\textbf{(left) IsoFLOPs curves.} For a fixed FLOPs budget, we train different models for the corresponding number of iterations with a matched cosine cycle length. For each isoFLOPs curve, we find a local minimum. \textbf{(middle)} We plot the model at the local minimum of each isoFLOPs curve and show the power law estimation. \textbf{(right)} We display our prediction with the ones from Hoffman and Kaplan. Our trained models are sampled in blue, while the range of models trained by \cite{kaplan2020scalinglawsneurallanguage} is represented by the gray area and the one from \cite{hoffmann2022trainingcomputeoptimallargelanguage} in the green area.
  }
  \label{fig:isoflops}
\end{figure*}

\section{Results}
\label{sec:results}
Our results are threefold: \textit{first}, we conduct a search for a scaling law related to transformers applied to graphs. \textit{Second}, we perform a thorough ablation study on model size, adjacency matrix, and positional encoding. \textit{Third}, we train a new family of transformers with masking pre-training that achieves new SOTA results on standard physics datasets. Additional results, including correlation between training loss, All-Rollout RMSE, and FLOPs, are available in \autoref{appendix:results}.

\subsection{Scaling Laws}
\label{subsec:scaling}

We start by investigating the following problem: \emph{given a fixed FLOPs budget, how should one balance the number of training steps and the size of its model?} Similar to \cite{hoffmann2022trainingcomputeoptimallargelanguage}, we model the final All-Rollout RMSE on the validation set as a function of $P$ the number of parameters and $D$ the number of training nodes.

For six given FLOPs budgets (see \autoref{annex:flops} for FLOPs computation), we train an extensive range of models\footnote{Models trained can be seen in \autoref{appendix:models}} for different number of training steps on the \dataset{3D-CoarseAneurysm} dataset and select the final training loss (a similar study yielding close results on the \dataset{Cylinder} dataset is available in \autoref{appendix:isoflops}). Models were configured with varying number of layers and embedding sizes.

Each training follows a cosine decay schedule matching the number of training steps the model must get to ensure fair comparisons between every run. For each FLOPs budget, we plot the parameters of the local minima model and fit a power law (see \autoref{fig:isoflops}). We find that $N \propto C^{0.75}$ with $C$ the FLOPs budget. While this value is much higher than \cite{hoffmann2022trainingcomputeoptimallargelanguage}, it is in the range of results obtained by \cite{kaplan2020scalinglawsneurallanguage}.

We highlight in \autoref{fig:isoflops} the size of models used in both papers and ours. Both \cite{kaplan2020scalinglawsneurallanguage} (\textit{resp.} \cite{hoffmann2022trainingcomputeoptimallargelanguage}) pretrain Large Language Models (LLM), with models ranging between 300k and 700M parameters (\textit{resp.} between 100M and 10B)\footnote{We train models ranging between 50k to 50M parameters}. They also fit power laws, finding $N \propto C^{0.73}$ (\textit{resp.} $N \propto C^{0.50}$). 

First, it shows that even on a very different task and domain, transformers follow very similar power laws. Second, this is evidence of a curvature in such power law, with the exponent decreasing in value for larger models. Based on said power law, the optimal size for a model trained on the \dataset{3D-CoarseAneurysm} for 30 epochs is around 50M parameters. We confirm this hypothesis by training a large base model (\textbf{XL}) of this size.

\subsection{Ablation Study}
\label{subsec:ablation}

\begin{figure*}[tbh!]
  \centering
  \includegraphics[width=1\textwidth]{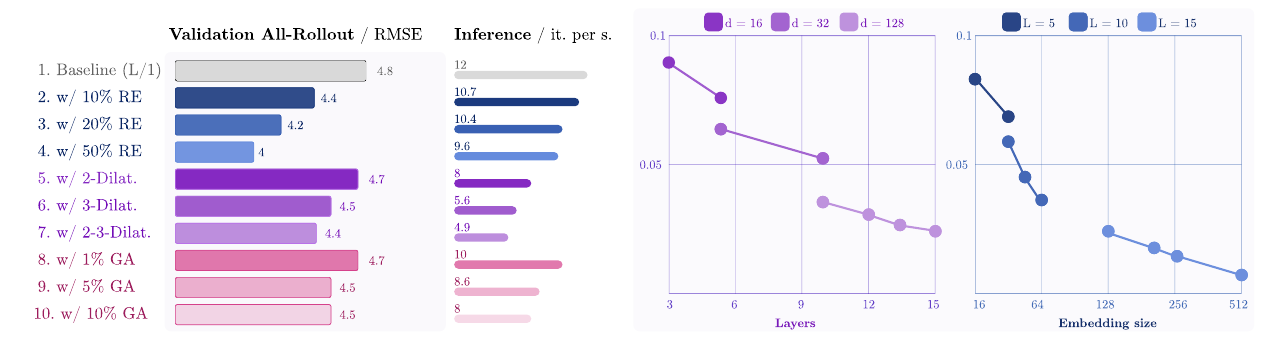}
  \caption{\small\textbf{(left)} All-Rollout RMSE for every $\mathbf{A}$ augmentations. \textbf{(right)} Final training loss with varying number of layers $L$ and embedding size $d$.
  }
  \label{fig:ablations}
\end{figure*}

\paragraph{Models and $K$-hop size}
We trained multiple models with different $K$-hop sizes, embedding sizes and number of processing layers in \autoref{fig:overview} and \autoref{fig:ablations}. In all cases, we find that increasing the model size or the number of neighbors leads to a smaller training loss and a smaller All-Rollout RMSE. It is important to note that an increase in the $K$-hop size leads to a large increase in vRAM usage and training time, due to the increase of non null values in $\mathbf{A}^K$. Overall, we find that directly increasing the $K$-hop size is often not worth it in terms of gain versus training time. However, it is much more interesting to increase either the width of our transformers or the number of layers (see \autoref{fig:ablations}).

\textbf{Random Edges (RE)} Adjacency matrix with random connections offers the best improvements for the smallest cost (15\% improvements for a 15\% longer training time). We add 20\% of random symmetric edges in our augmented adjacency matrix.

\textbf{Dilation (Dilat.)} We find that using a \textit{Dilation-2} to simulate larger receptive fields at a fraction of the cost of increasing the hop size is particularly effective (see \autoref{fig:ablations}). For larger dilation, we find that the improvements are not worth the training and inference time increase. We use the \textit{Dilation-2} in our augmented adjacency matrix.

\textbf{Global Attention (GA)} Adding global attention leads to a 10\% improvement with an increase in training time of up to 30\%. We use 1\% of global nodes in our augmented adjacency matrix. For the rest of the training, we always use an augmented adjacency matrix with 20\% random edges, \textit{Dilation-2}, and 1\% of global nodes. An \textbf{M/1} model trained with said matrix outperforms our \textbf{M/2} model while being twice as fast.

\textbf{Positional Encoding}
Using Laplacian Position Encoding or RandomWalk Position Encoding leads to poorer results than using the 3D geometrical coordinates. 
A model without any sort of positional encoding performs worse for our CFD datasets: an \textbf{M/1} model trained without 3D-coordinates performs similar to an \textbf{S/1} model. 

Overall, all adjacency matrix augmentations improve the model’s performance, but at the cost of slower training. Random edges is the augmentation offering the best trade-off. For Global Attention, their effectiveness appears to depend on the dataset: some datasets benefit more from this augmentation, while others perform better using global features at each node.

\subsection{Results}
\label{subsec:results}
Our largest model (\textbf{XL/1}) with a regular adjacency matrix and no pre-training already outperforms the current SOTA by $25$\%. When augmented with an Adjacency Matrix and pre-trained using masking, it achieves an average improvement of $38.8$\% over the current SOTA. Additionally, our \textbf{L/1} model, even without any augmentation, improves on MeshGraphNet by $30$\% while being twice as fast. While our XL/1 model has a much higher number of parameters than previous architectures, it has similar training speed and memory consumption as shown in \autoref{fig:overview}, which justifies the comparison. The figure also showcases standard-deviation over 5 runs.

A detailed view of the attention mechanism on meshes can be seen in \autoref{appendix:attentionvizu}.
Even our smallest model \textbf{S/1} (see \autoref{fig:overview}) achieves similar results to MeshGraphNet while being $7\times$ faster, without any pre-training or augmentation. Detailed results can be seen in \autoref{tab:main-res}. It is important to note that such scaling cannot be achieved with an MPS architecture, as already with 15 layers one reaches limits in both training time and vRAM capacities on a single A100 GPU, mostly due to the gradient back-propagation on the edges.

Our \textbf{XL} model achieves a $15\times$ speed-up over classical CFD methods on the \dataset{Cylinder} dataset and a $500\times$ speed-up on the \dataset{3D-Aneurysm} dataset (see \autoref{tab:datasets-details}). Even accounting for the model's training time and dataset construction time, one would need to simulate only 120 aneurysms before our machine learning approach becomes more time-efficient (see \autoref{fig:datasetsdetails-time}).

\textbf{Out-of-distribution meshes} We evaluated the performance of our model on out-of-distribution meshes by training a model on coarse aneurysms before testing it on fine aneurysms (10k to 250k nodes). 
Even with such a gap, our model performs only 80\% worse than a model trained on fine meshes, which remains better than results from MGN trained on the fine meshes.

\textbf{Limitations on the \dataset{Plate} dataset} This is the only dataset where our architecture yields only marginal improvements over MGN and doesn't beat the SOTA. We attribute this to a number of reasons. Since only a small fraction of the second object’s nodes are actually relevant (the ones being in contact with the fixed plate), multigrid approaches are very efficient because they can select only these relevant nodes and keep more computing power for the fixed plate. Additionally, our model primarily learns local interactions between adjacent mesh nodes, making it much harder to successfully model interactions that emerge from world edges, as shown in the appendix by the lack of attention to world edges. Moreover, by removing the edge’s features, we remove a powerful indicator of its type (\textit{i.e.}, natural or artificial).

We believe this new family of models provides a powerful and efficient method for simulating complex physics on very large meshes. Leveraging a simple and well-established architecture, combined with straightforward augmentations of the Adjacency Matrix and appropriately tuned training steps, our models surpass previous architectures in performance while being significantly faster. For instance, they can accurately predict blood flow within an aneurysm, offering valuable insights for critical medical metrics (see \autoref{subappendix:aneurysms}).

\begin{table*}[!th]
  \centering
  \caption{\label{tab:main-res}All numbers are $\times 10^{-3}$. \dataset{Dataset}-1 means one-step RMSE, and \dataset{Dataset}-All means all-rollout RMSE. MeshGraphNet (MGN) results are reproduced according to \cite{pfaff2021learning}, BSMS-GNN according to \cite{cao2023efficientlearningmeshbasedphysical} and
  Multigrid and Masking according to \cite{garnier2024meshmask}. Best results are in \color{red}\textbf{Red}\color{black}, second in \textbf{bold}.}
  \resizebox{0.99\linewidth}{!}
  {
    \begin{tabular}{@{}lllllll@{}}
      \toprule
      \sc{Model}   & \sc{Cylinder} & \sc{Plate} & \sc{Airfoil} &  \sc{2D-Aneurysm}  &  \sc{3D-CoarseAneurysm} &  \sc{3D-Aneurysm} \\
                                                        & \sc{1-rmse $\downarrow$} & \sc{1-rmse $\downarrow$} &
                                                        \sc{1-rmse $\downarrow$} &
                                                        \sc{1-rmse $\downarrow$} &
                                                        \sc{1-rmse $\downarrow$} & \sc{1-rmse $\downarrow$} \\
                                        \midrule
MGN  & 2.52   & \color{red}\textbf{0.07} & 329 & 794 & 1420 & 1795 \\
\cite{lino2021simulating} & 2.7   & \textbf{0.10} & \textbf{300} & - & -  & - \\ 
BSMS-GNN & 2.83   & 0.15 & 314 & \textbf{632} & 1137  & \textbf{719} \\
GATv2 & 2.7    & 0.16   & - & -    & 1622     & -  \\ 
TransformerConv & 2.68    & 0.18   & - & -    & 2177     & -  \\ 
Multigrid & 2.9   & 0.17 & 302 & 638 & -  & 749 \\ 
Masked Multigrid  & \textbf{2.5} & 0.11 & 310 & 645 & \textbf{692} & 725 \\ 
\midrule
Masked XL/1 & \color{red}\textbf{2.3}  & 0.2 & \color{red}\textbf{289} & \color{red}\textbf{419.4}  & \color{red}\textbf{340.8}  & \color{red}\textbf{395.8} \\            
\bottomrule
\bottomrule
& \sc{all-rmse $\downarrow$} & \sc{all-rmse $\downarrow$} & \sc{all-rmse $\downarrow$} & \sc{all-rmse $\downarrow$}  & \sc{all-rmse $\downarrow$} & \sc{all-rmse $\downarrow$} \\ 
\midrule
MGN  & 46.9   & 16.9   & 11398 & 7513   & 7648   & 13747    \\
\cite{lino2021simulating} & 61.2    & 15.7  & 10272 & -    & -     & -  \\ 
BSMS-GNN & 60.5    & 16.6   & 9418 & 6983    & 7198     & 10993  \\ 
GATv2 & 55.7    & 17   & - & -    & 8391     & -  \\ 
TransformerConv & 52.1    & 17.3   & - & -    & 9420     & -  \\
Multigrid & 56.9    & \textbf{8.1}   & 9871 & 7009    & -     & 11327  \\ 
Masked Multigrid & \textbf{29}  & \color{red}\textbf{4.5}  & \textbf{8794} & \textbf{6489}    & \textbf{6421}   & \textbf{8772}  \\ 
\midrule
Masked XL/1  & \color{red}\textbf{13.5} & 13.8  & \color{red}\textbf{6453.5} & \color{red}\textbf{3316.2} & \color{red}\textbf{2428.4}  & \color{red}\textbf{4825.1} \\
 \bottomrule
\end{tabular}
}
\label{table:all-results}
\end{table*}

\section{Conclusion and Limitations}
\label{sec:conclusion}

In this work, we introduced a novel Transformer-based GNN architecture specifically designed for physics-based simulations, utilizing the adjacency matrix as a direct attention mask within a Transformer framework. 
By augmenting the adjacency matrix with Dilated Sliding Windows, Global Attention, and Random Connections, our model effectively captures long-range dependencies with improved efficiency. 
Through comprehensive scaling law analysis and ablation studies, we identified optimal configurations for model size and receptive fields, showcasing the scalability and robustness of our approach. 
Collectively, these advancements push the boundaries of state-of-the-art physics simulations using graph-based neural networks. As mentioned above, our model does not perform as well (while still better than MGN) on the \dataset{Plate} Dataset. 
We also found evidence that the gap with other models is reduced when fewer features per node are available. Another downside is that we removed the edge’s features and parsed the node’s position directly as features. 
As a result, our architecture isn’t invariant to geometric changes, whether translation, rotation, scaling, or other augmentations. Addressing these limitations will be the focus of future work.

\begin{ack}
The authors acknowledge the financial support from ERC grant no 2021-CoG-101045042, CURE. Views and opinions expressed are however those of the author(s) only and do not necessarily reflect those of the European Union or the European Research Council. Neither the European Union nor the granting authority can be held responsible for them.

The authors thank Arthur Verrez and Loïc Chadoutaud for valuable feedback on the manuscript. 
\end{ack}

\newpage

\bibliography{example_paper}
\bibliographystyle{icml2025}


\appendix

\section{Datasets}
\label{appendix:datasets}

\begin{figure}[!t]
  \centering
  \includegraphics[width=0.99\textwidth]{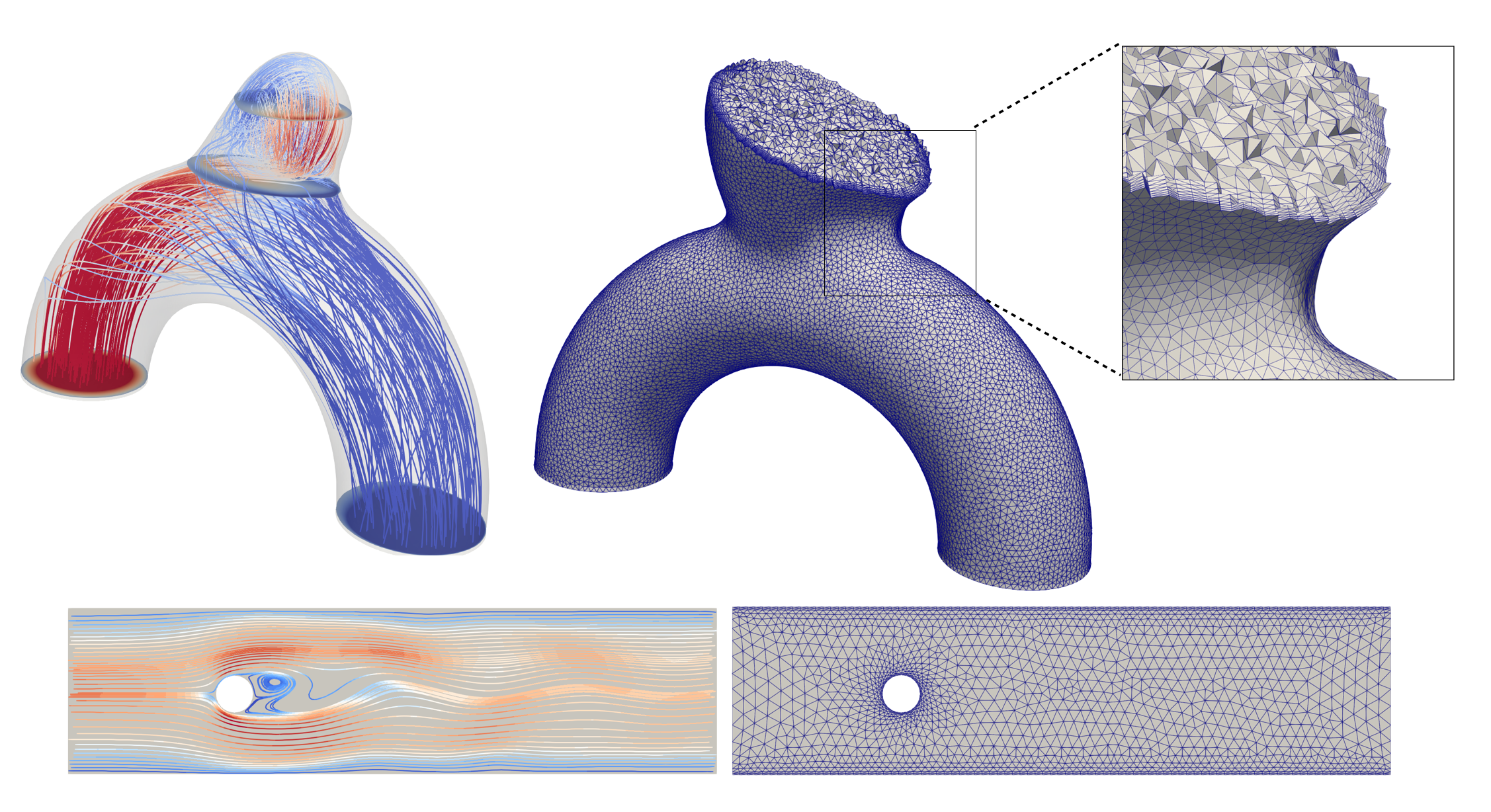}
\caption{
We display details of 2 datasets: \dataset{3D-Aneurysm} and \dataset{Cylinder}. \textbf{(top)} We display the Systolic velocity streamlines colour-coded with the vertical component on the left, and a detail overview of the mesh of the right. \textbf{(bottom)} We display streamlines of the velocity around a cylinder on the left, and a view of the mesh used for simulation on the right.
} 
  \label{fig:datasetsdetails}
\end{figure}

We give details below about the inputs and outputs used for each dataset (see Table \ref{tab:datasets-details} and \autoref{fig:datasetsdetails}). \dataset{Cylinder}, \dataset{Plate} were generated with COMSOL \cite{comsol} and were introduced by \cite{pfaff2021learning}.
\dataset{Airfoil} was generated with SU2 \cite{Economon2016} and was introduced by \cite{pfaff2021learning}. \dataset{3D-Aneurysm} was generated with CimLib \cite{cimlib} and was introduced by \cite{aneurysmdataset}.

\dataset{2D-Aneurysm} was generated by re-meshing slices from the original \dataset{3D-Aneurysm} dataset. \dataset{3D-CoarseAneurysm} was generated by coarsening the meshes from \dataset{3D-Aneurysm} before interpolating the velocity fields onto it.

\renewcommand{\arraystretch}{1.1}%
\begin{center}
\begin{small}
\begin{tabular}[p]{|p{35.5mm}||c|c|c|c|c|c|c|c|c|}
	\hline
	\textbf{Dataset} &  
	\textbf{Inputs} &
	\textbf{Outputs} &
        \textbf{History} &
        \textbf{$t_{gnn}$ (ms/step)} &
        \textbf{$t_{gt}$ (ms/step)} 
  \\\hline\hline
    \dataset{Cylinder} & $n, v_x, v_y$ & $v_x, v_y$ & 0 & 49.3 & 820 \\\hline
    \dataset{Plate} & $n, x, y, z, f_{\text{in}}$ & $x, y, z, \sigma$ & 0 & 60.4 & 2893 \\\hline
    \dataset{Airfoil} & $n, v_x, v_y, \rho$ & $v_x, v_y, \rho$ & 0 & 68 & 11015 \\\hline
    \dataset{2D-Aneurysm} & $n, v_x, v_y, v_{\text{in}}$ & $v_x, v_y$ & 1 & 172 & - \\\hline
    \dataset{3D-CoarseAneurysm} & $n, v_x, v_y, v_z, v_{\text{in}}$ & $v_x, v_y, v_z$ & 1 & 125 & - \\\hline
    \dataset{3D-Aneurysm} & $n, v_x, v_y, v_z, v_{\text{in}}$ & $v_x, v_y, v_z$ & 1 & 942 & 540000 \\\hline
\end{tabular}
\label{tab:datasets-details}
\end{small}
\end{center}
\renewcommand{\arraystretch}{1.0}%

In Table \ref{tab:datasets-details}, $n$ is the node type (Inflow, Outflow, Wall, Obstacle, Normal) and $v_{\text{in}}$ the inflow velocity at the current timestep. When history is different than $0$, we use a first-order derivative of the inputs as an extra feature. For example, we add $a_x, a_y, a_z$ to each node from an aneurysm mesh.

\begin{figure}[!t]
  \centering
  \includegraphics[width=0.99\textwidth]{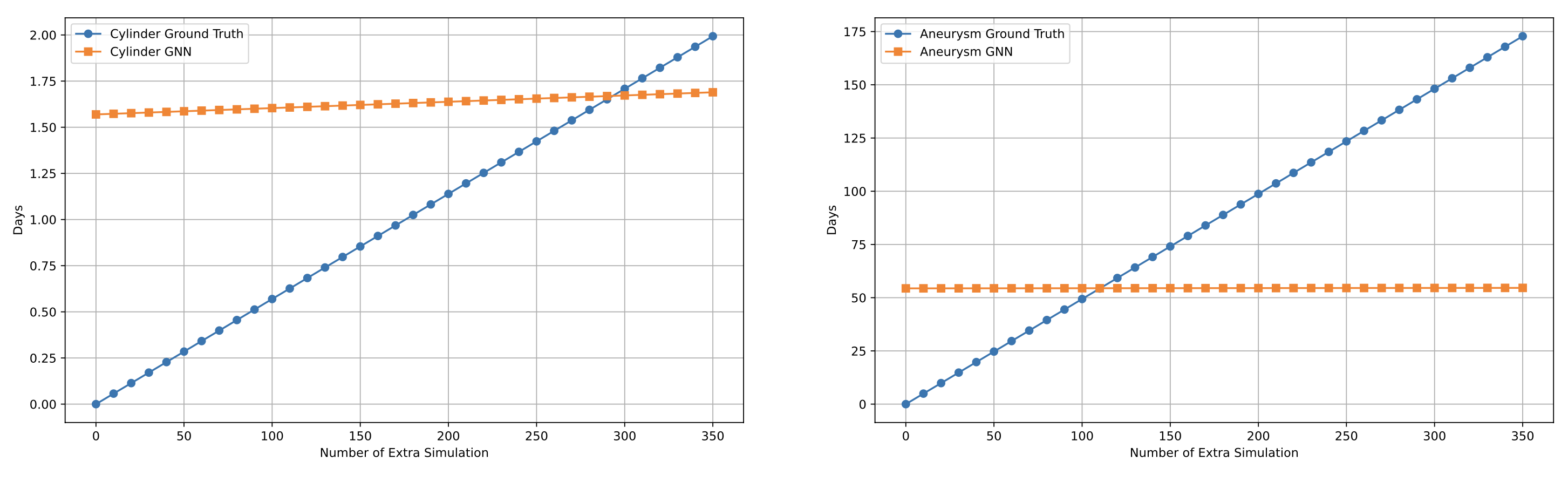}
\caption{
Time spent in days for the training of a \textbf{XL/1} transformer on Cylinder and Aneurysm followed by inference simulations. For the groundtruth, we consider time from the \cite{cimlib} solver. For the GNN, we consider the inference time, the training time, as well as the duration from the groundtruth solver to build the datasets. We can see that even when taking everything into account, only 100s of simulations are necessary for the machine learning approach to be worth it.
} 
  \label{fig:datasetsdetails-time}
\end{figure}

\subsection{Noise and Global Attention}

We display the noise used as well as the global nodes selected for each dataset in Table \ref{tab:noises}.

\paragraph{Noise Scale} We make our inputs noisy by following the same strategy as \cite{sanchezgonzalez2020learning}.
We add random noise $\mathcal{N}(0,\sigma)$ to the dynamical inputs. Each noise standard deviation was either selected from previous papers (\dataset{Cylinder}, \dataset{Plate} and \dataset{Airfoil}) or selected by looking at average one-step error in predictions (\dataset{2D-Aneurysm}, \dataset{3D-CoarseAneurysm} and \dataset{3D-Aneurysm}). More precisely, we train one model without any noise and then compute the distribution of the one-step error. For the case of the aneurysm dataset, the variational nature of the inflow makes for said distribution to be time-dependant. While we investigated different strategies, such as time-dependant noise, we simply used the larger standard deviation possible($\max\limits_{t \in [1, T-1]}\sigma _t$). 

\paragraph{Global Attention}  We also display which nodes were considered for the Global Attention. Wall Nodes represent both the nodes on the top and bottom wall and the cylinder. Inlet Nodes represent nodes at the inflow boundary in the artery. We selected those nodes based on the amount of influence their boundary conditions have onto the FEM solver.

\renewcommand{\arraystretch}{1.1}%
\begin{center}
\begin{small}
\begin{tabular}[p]{|p{35.5mm}||c|c|c|c|c|c|c|}
	\hline
	\textbf{Dataset} &  
	\textbf{Noise} &
        \textbf{Global Attention}
  \\\hline\hline
    \dataset{Cylinder} & 0.02 & Wall Nodes \\\hline
    \dataset{Plate} & 0.003 & Obstacle Nodes \\\hline
    \dataset{Airfoil} & $v_x, v_y$: 10, $\rho$: 0.02 & Airfoil Nodes \\\hline
    \dataset{2D-Aneurysm} & 10 & Inlet Nodes \\\hline
    \dataset{3D-CoarseAneurysm} & $v_x, v_y$: 10, $v_z$: 0.5 & Inlet Nodes \\\hline
    \dataset{3D-Aneurysm} & $v_x, v_y$: 10, $v_z$: 0.5 & Inlet Nodes \\\hline
\end{tabular}
\label{tab:noises}
\end{small}
\end{center}
\renewcommand{\arraystretch}{1.0}%

\section{Models}
\label{appendix:models}

\begin{figure}
  \centering
  \includegraphics[width=\columnwidth]{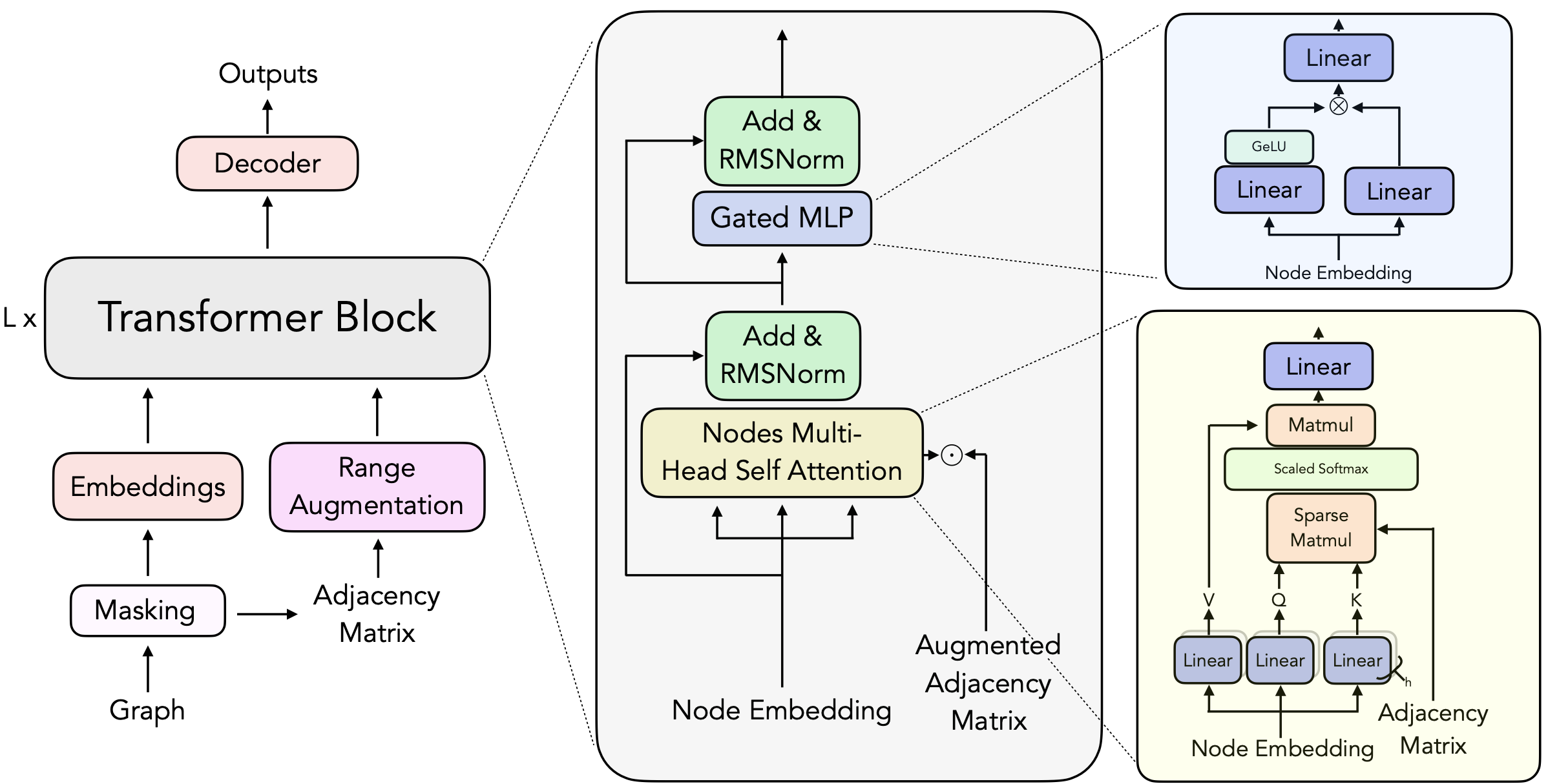}
  \caption{\small\textbf{The Masked Transformer architecture.} \textbf{(left)} Our model takes the graph nodes' features and the adjacency matrix as inputs. The adjacency matrix is improved with Dilation, Global Attention, and Random Jumpers. \textbf{(middle)} Each transformer is made of masked Multi-Head Self-Attention followed by a Gated MLP with residual connection and Layer Normalization. \textbf{(right)} The Multi-Head Self-Attention masks the node's features by computing a sparse matrix multiplication indexed on the Augmented Adjacency matrix. The Gated MLP processes the features in two branches with an expansion factor $e$.}
  \label{fig:archi}
\end{figure}

We list all the models trained in \autoref{tab:all_models}, with $d$ the embedding size, gated size the input size of the Gated MLP, $h$ the number of attention heads and $L$ the number of layers. Note that each model may have been trained multiple times for different optimizers, learning rate schedule, and training steps.

\begin{longtable}[h!]
{r | r |rrrr }
\toprule
Model & Parameters (million) &  $d$ &  gated size &   $h$ &  $L$ \\
\midrule
    & 0,012 & 16 & 48 & 2 & 3 \\
    & 0,019 & 16 & 48 & 2 & 5 \\
    & 0,072 & 32 & 96 & 2 & 5 \\
    & 0,14 & 32 & 96 & 2 & 10 \\
    & 0,31 & 48 & 144 & 2 & 10 \\
    \textbf{S} & 0,55 & 64 & 192 & 2 & 10 \\
    & 0,85 & 80 & 240 & 2 & 10 \\
    & 2,18 & 128 & 384 & 4 & 10 \\
    & 2,6 & 128 & 384 & 4 & 12 \\
    & 3 & 128 & 384 & 4 & 14 \\
    \textbf{M} & 3,2 & 128 & 384 & 4 & 15 \\
    & 4,5 & 152 & 456 & 4 & 15 \\
    & 8 & 200 & 600 & 4 & 15 \\
    \textbf{L} & 13 & 256 & 769 & 4 & 15 \\
   \textbf{XL} & 51 & 512 & 1536 & 4 & 15 \\
 
\bottomrule
\caption{\textbf{All models.}
Hyperparameters and size of all models trained for the ablation studies and isoFLOPs curves.}
\label{tab:all_models}
\end{longtable}

\section{Theoretical Considerations}
\label{annex:th-consideration}

 It’s important to highlight that a significant part of our improvements is provided by better data management. Our architecture enables the model to process the same amount of information more efficiently, improving flow and allowing for a much larger number of parameters. This design directly mitigates the over-squashing problem commonly seen in GNNs.

In addition, our adjacency matrix augmentations function similarly to a multigrid method, facilitating faster information propagation across the graph compared to standard message-passing approaches.
Building on this, and following the insights from \cite{veličković2018graph}, \cite{kim2022puretransformerspowerfulgraph}, and \cite{brody2022attentivegraphattentionnetworks}, we now examine more closely why our architecture may outperform traditional message-passing methods.

A standard 1-hop GNN updates node features by aggregating from immediate neighbors:

\begin{align}
m_v^{(\ell)} &= \operatorname{MES}^{(\ell)}\Big( \{ (h_u^{(\ell-1)},e_{uv}) : u\in Q_1(v,G) \} \Big), \label{eq:1hop_mes}\\[1mm]
h_v^{(\ell)} &= \operatorname{UPD}^{(\ell)}\Big( m_v^{(\ell)}, h_v^{(\ell-1)} \Big), \quad \ell=1,\dots,L. \label{eq:1hop_upd}
\end{align}

In comparison, one could summarize the attention as follows:

\begin{align}
\mathbf{z’}_i^{(\ell+1)} 
\;=\; 
\sum_{j \in \mathcal{N}(i)} 
\alpha_{ij}^{(\ell)}\, 
\bigl(W^V \mathbf{z}_j^{(\ell)}\bigr), 
\quad 
\alpha_{ij}^{(\ell)} 
\;=\; 
\frac{\exp\!\Big(\bigl(W^Q\mathbf{z}_i^{(\ell)}\bigr)^\top\bigl(W^K\mathbf{z}_j^{(\ell)}\bigr)\Big)}{\sum_{k \in \mathcal{N}(i)} 
\exp\!\Big(\bigl(W^Q\mathbf{z}_i^{(\ell)}\bigr)^\top\bigl(W^K\mathbf{z}_k^{(\ell)}\bigr)\Big)}.
\end{align}
\begin{align}
\mathbf{z}_i^{(\ell+1)}
\;=\; W^o\mathbf{z'}_i^{(\ell+1)}
\end{align}

We can also assume that the Feed Forward Layer from the Transformer and the Update function (after the edge’s information is transferred) actually serve the same purpose and work similarly. Thus, the question becomes: How is information traveling within the Aggregate function and the Attention layers? We believe this comes down to 2 main differences:
\begin{enumerate}
    \item in message passing, each edge is processed independently as a large batch of tokens
    \item in message passing, information is squashed using an aggregation function 
\end{enumerate}

Instead of a single aggregator (e.g.\ sum/mean) or a small MLP that outputs attention coefficients, we have multiple heads plus the $\mathrm{softmax}\bigl(\tfrac{QK^T}{\sqrt{d}}\bigr)$.
This allows richer, context‐dependent weighting of neighbors since each head can highlight different features or patterns in the neighbor set. 

This weighting is also one reason why a transformer-based model suffers much less from over-squashing: the softmax on $\langle Q_v, K_u\rangle$ can “turn off” irrelevant neighbors or strongly amplify crucial ones. This can preserve feature diversity across layers, similar to an anisotropic relaxation operator. 

Finally, an analysis of gradient backpropagation can show that attention yields better gradient signals for each neighbor rather than a rigid one due to the aggregation function in message passing.

\section{FLOPs Computation}
\label{annex:flops}

Below, we detail the computations to obtain the number of FLOPs per model for our architecture and a Message-passing network. We use a factor of $2$ for the Multiply-Accumulate cost. See \cite{flopstransformers} for a good introduction to FLOP computing.

\paragraph{Transformer}

\begin{itemize}
    \item Encoder: $2\times \text{input\_size} \times d \ll d^2$ 
    \item $QKV$: $3 \times 2 \times d^2$
    \item Scaled Softmax and Query reduction $\ll d^2$ 
    \item Projection: $2 \times d^2$
    \item Gated MLP: $2\times \text{expanded\_branches} + \text{hadamard} + \text{final\_branch} = 2 \times 6d^2 + O(d^2) + 6d^2 = 18d^2$
    \item Decoder: $2\times d \times \text{output\_size} \ll d^2$ 
\end{itemize}

For a total of FLOPs $\approx L \times 26d^2$

\paragraph{Message Passing}

\begin{itemize}
    \item Encoder: $2\times \text{input\_size} \times d + 3 \times 2 \times d^2$ 
    \item Edge Block: $2 \times 3d \times d + 3 \times 2 \times d^2$
    \item Node Block: $2 \times 2d \times d + 3 \times 2 \times d^2$
    \item Decoder: $3 \times 2 \times d^2 + 2\times d \times \text{output\_size}$ 
\end{itemize}

For a total of FLOPs $\approx 6d^2 + L \times 22d^2$

We follow the process from \cite{kaplan2020scalinglawsneurallanguage} and consider that the backward pass has twice the FLOPs of the forward pass for the training process. For the model FLOPs we compare their approximation of FLOPs $\approx 2 \times \text{\# Parameters}$ to our method. Due to our Gated MLP, our method is much closer to 2P than the usual $24d^2$ of standard transformers. Thanks to this, we use the approximation FLOPs $\approx 2P$ for all our computations in the paper. On the other hand, this estimation grossly overestimates the FLOPs of a Message Passing Network. We do not use it when computing the FLOPs of such model.

\begin{table*}[h!]
\centering
\begin{tabular}{c c c c | c | c}
\toprule
Parameters (million) &  $d$ &  gated size &  $L$ & FLOP Ratio (Us/$2P$) & (MPS/$2P$)\\ 
\midrule
0.550 & 64 & 192 & 10 & 0.97 & 0.84\\
3.2 & 128 & 384 & 15 & 0.99 & 0.86\\
13 & 256 & 768 & 15 & 0.98 & 0.85\\
52 & 512 & 1536 & 15 & 0.98 & 0.85\\
\bottomrule
\end{tabular}
    \caption{\textbf{FLOP comparison.} 
    For various model sizes, we show the ratio of the FLOPs that we compute per sequence to those using the $2P$ approximation. We also show that the $2P$ approximation overestimates the FLOPs of a Message Passing Network.
    }
    \label{tab:flops}
\end{table*}

\section{Results}
\label{appendix:results}

\subsection{Metrics}
\label{appendix:metrics}
To evaluate our models, we use the 1-step RMSE and the All-Rollout RMSE defined below:

\begin{align*}
    \text{1-step}(f) := \frac{1}{TN} \sum _{t=1}^{T} \Bigg(\sum _{i \in V} \big(G_t - f(G_{t-1})\big)_i^2 \Bigg)
\end{align*}

\begin{align*}
    \text{All-Rollout}(f) := \frac{1}{TN} \sum _{t=1}^{T} \Bigg(\sum _{i \in V} \big(G_t - \underbrace{f \circ ... \circ f}_\text{t times}(G_0)\big)_i^2 \Bigg)
\end{align*}

where $f$ is the model, $T$ the number of time steps, $N$ the number of nodes and $G_t$ the ground truth graph at time step $t$.

\subsection{IsoFlops on Cylinder}
\label{appendix:isoflops}

\begin{figure}[h!]
  \centering
  \includegraphics[width=0.99\textwidth]{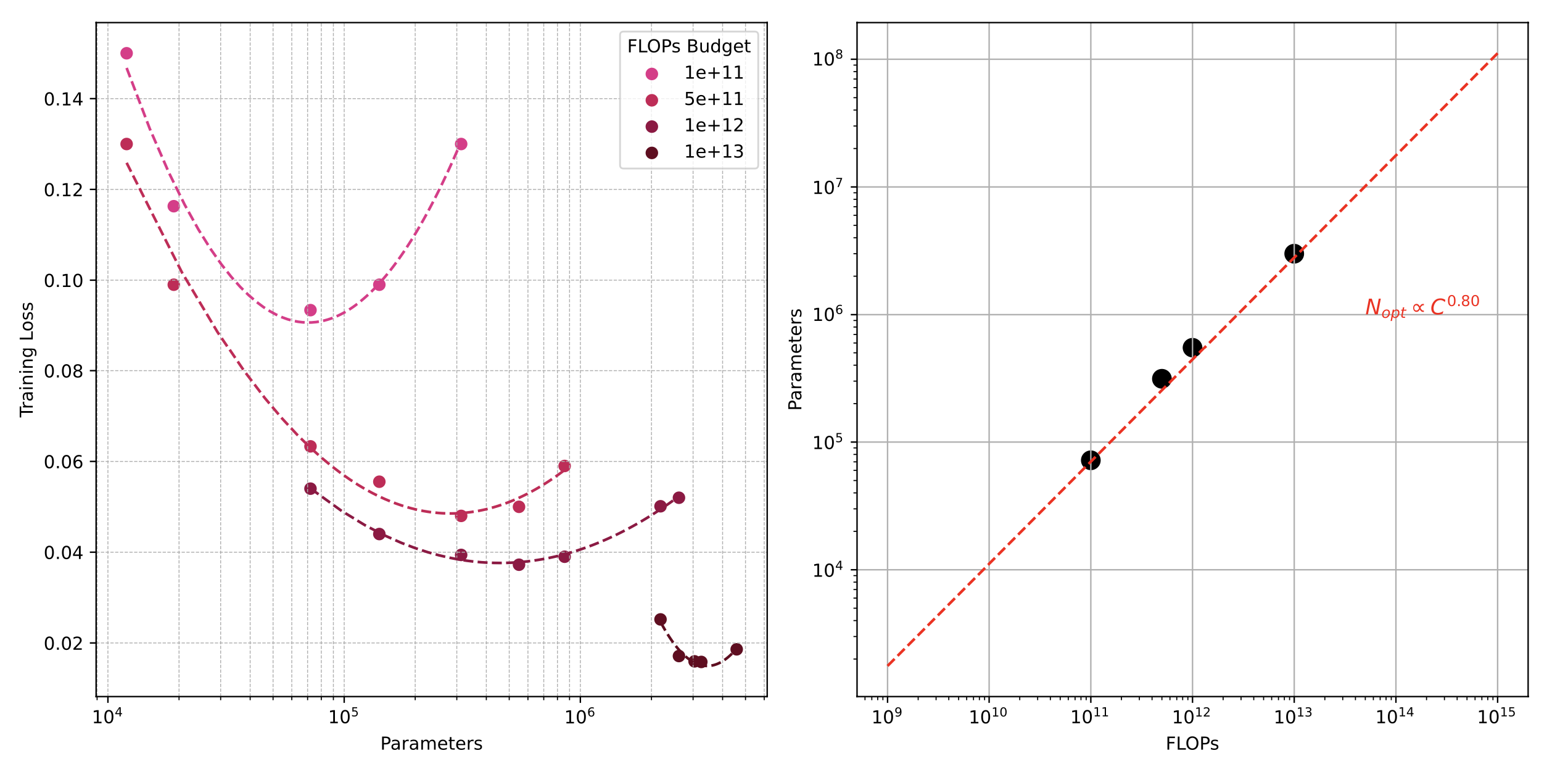}
  \caption{\textbf{(left) IsoFLOPs curves.} For a fixed FLOPs budget, we train different models for the corresponding number of iterations with a matched cosine cycle length. For each isoFLOPs curve, we find a local minimum. \textbf{(right)} We plot the model at the local minimum of each isoFLOPs curve and show the power law estimation.}
  \label{fig:isoflopscylinder}
\end{figure}

We show the result of the same isoFLOP search as described in the main paper but on the \dataset{Cylinder} dataset. Results are shown in \autoref{fig:isoflopscylinder}. We conducted this experiment with a smaller range of both FLOP and model size, and with fewer tokens (1500 instead of 10000) per iteration. 

We find the scaling to be very similar, although we find a slightly bigger exponent factor. This is in line with a potential curvature in the scaling laws.

\subsection{Loss and RMSE correlation}
\label{subappendix:correlation}

\begin{figure}[h!]
  \centering
  \includegraphics[width=0.99\textwidth]{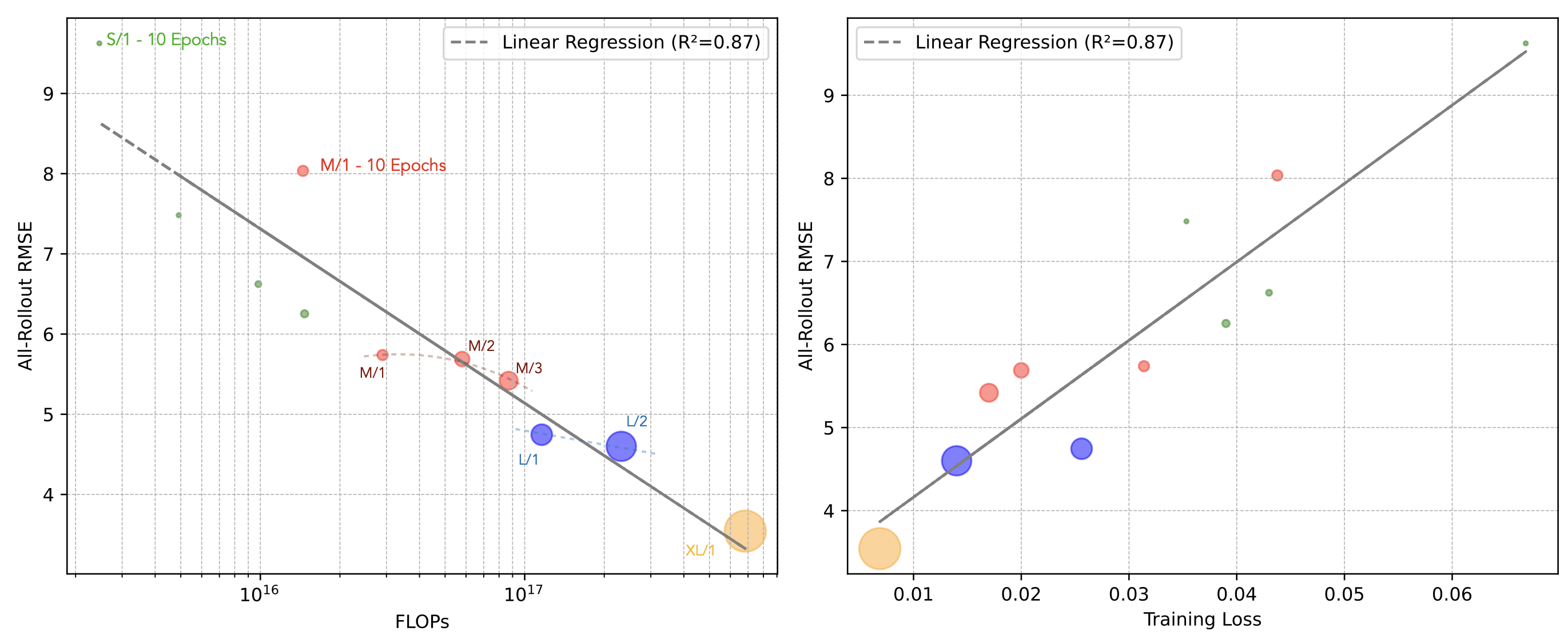}
  \caption{\small\textbf{(left)} We plot the FLOPs versus the All-Rollout RMSE on the testing set and find a strong correlation. \textbf{(right)} We plot the training loss versus the All-Rollout RMSE on the testing set and find a strong correlation.}
  \label{fig:losscorrelation}
\end{figure}

In \autoref{fig:losscorrelation}, we first show that the FLOPs and the All-Rollout RMSE are strongly correlated. We find that scaled-up models are more efficient and keep improving the evaluation metric. We also show a strong correlation between the training loss and the evaluation metric. While the correlation between the training loss and the 1-Step RMSE can be easy to see (given no over-fitting), it is much less obvious for the All-Rollout RMSE. This shows how powerful adding noise is to mitigate error propagation.

\subsection{Learning rate and Optimizers}
\label{subappendix:other-differences}

\begin{figure}[h!]
  \centering
  \includegraphics[width=0.99\textwidth]{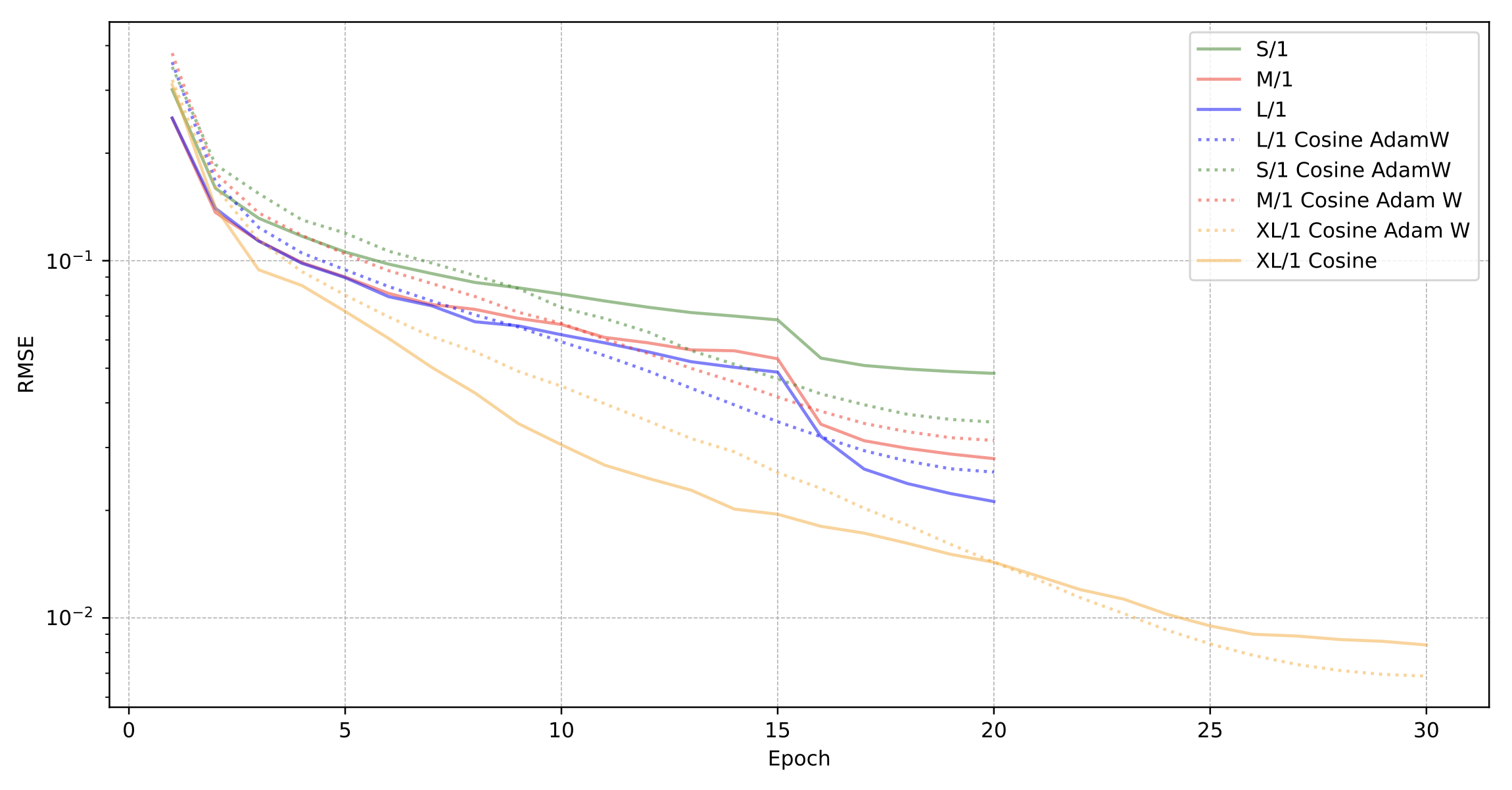}
  \caption{\small\textbf{Optimizer and Learning Rate.} We plot training loss for models trained with Adam, AdamW, an exponential decay schedule, and a cosine schedule. The cosine schedule and AdamW make All-Rollout RMSE more stable across training and provide better results given sufficient training time.}
  \label{fig:adamw}
\end{figure}

We showcase in \autoref{fig:adamw} differences in training between using an Adam \cite{kingma2017adam} optimizer and an AdamW \cite{loshchilov2019decoupledweightdecayregularization} optimizer. We also show the differences between using different learning rate schedules: $10^{-4}$ for 75\% of the steps followed by an exponential decay to $10^{-6}$ for the remaining steps, versus a learning rate with warmup and cosine decay from $10^{-3}$ to $10^{-6}$ for the S, M and L models and from $10^{-4}$ to $10^{-7}$ for the XL model.

We find that a cosine decay schedule always outperforms the former strategy and, more importantly, that it makes the evaluation of All-Rollout RMSE much more stable. We also find that training with Adam decreases the training loss more aggressively but is ultimately outperformed by AdamW given sufficient training time.

\subsection{Comparison with Message Passing}
\label{subappendix:comparison}

\begin{figure}[h!]
  \centering
  \includegraphics[width=0.99\textwidth]{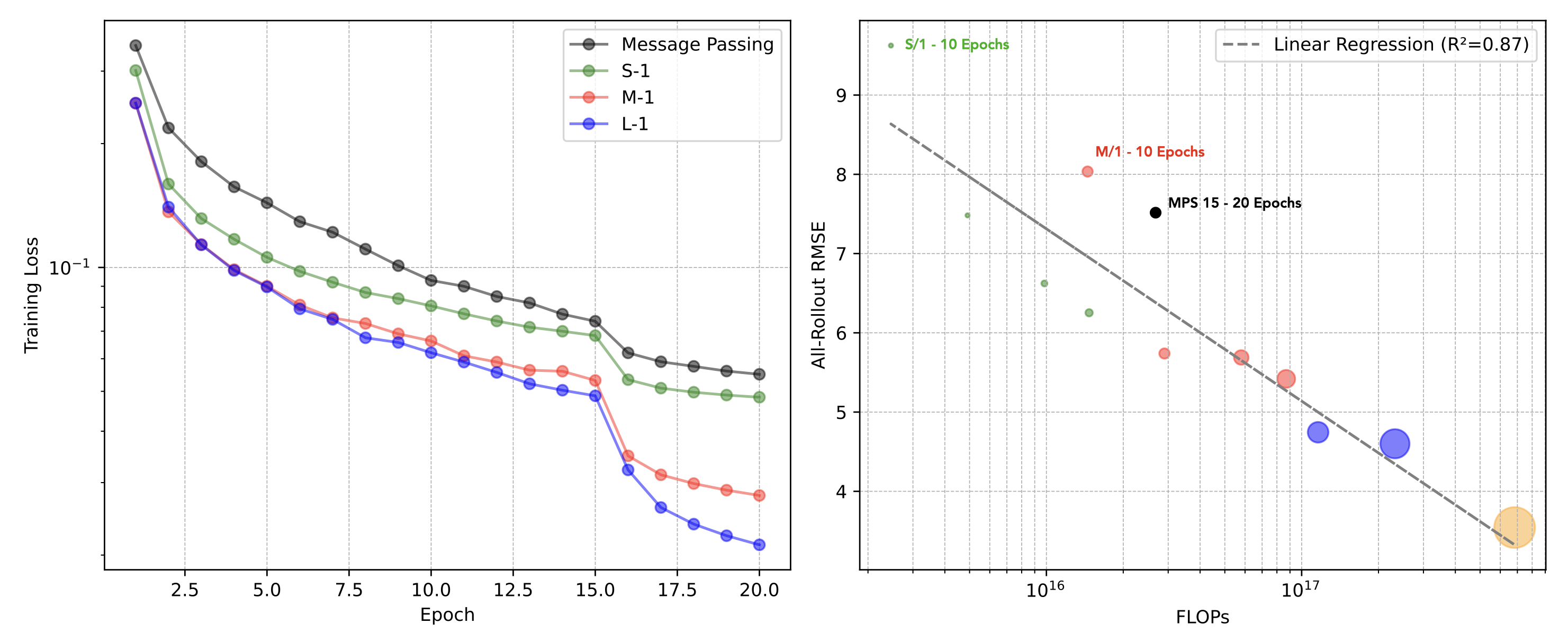}
  \caption{\small\textbf{(left)} We plot the training loss on 20 epochs of our models versus a message-passing model with $15$ layers. We find the loss to be higher than even our smallest model. \textbf{(right)} We find that the message-passing model makes less efficient training for the same FLOPs as our transformer models. Even trained for more epochs, it behaves like our under-trained transformers.}
  \label{fig:mpsloss}
\end{figure}

In \autoref{fig:mpsloss}, we show the differences between the best Message Passing Architecture and our transformer models during training. Even our smallest model achieves a better loss during the training, with $5.5\times$ fewer parameters.

We also show that the message-passing architecture is much less efficient FLOPs-wise than our models. For the same number of FLOPs, our \textbf{M/1} trained for the same number of epochs performs much better. It is also largely above our linear regression for FLOPs \textit{vs.} All-Rollout RMSE, ranging with our under-trained transformers. This shows that the message-passing architecture is much less efficient and needs more parameters and FLOPs to achieve performances similar to those of our transformers.

\subsection{Attention Vizualization}
\label{appendix:attentionvizu}

\begin{figure}[h!]
  \centering
  \includegraphics[width=0.99\textwidth]{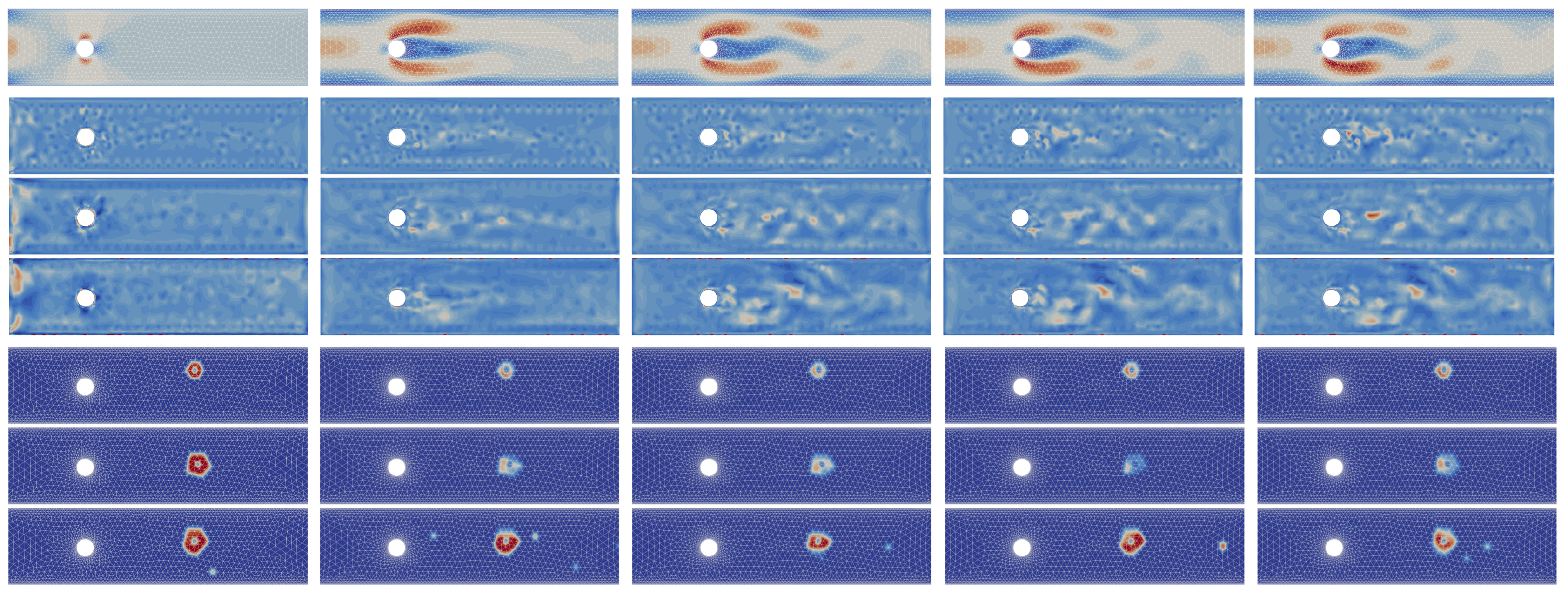}
  \caption{Fields are captured from $t=0$ and every $50$ time steps. \textbf{Row 1}: Velocity field with mesh. \textbf{Row 2, 3, 4} Magnitude of the attention for each node, in the order $A$, $A^2$ and Augmented $A$. \textbf{Row 5, 6, 7} Magnitude of the attention from a single node, in the order $A$, $A^2$ and Augmented $A$.}
  \label{fig:attention-cylinder}
\end{figure}

\begin{figure}[h!]
  \centering
  \includegraphics[width=0.60\textwidth]{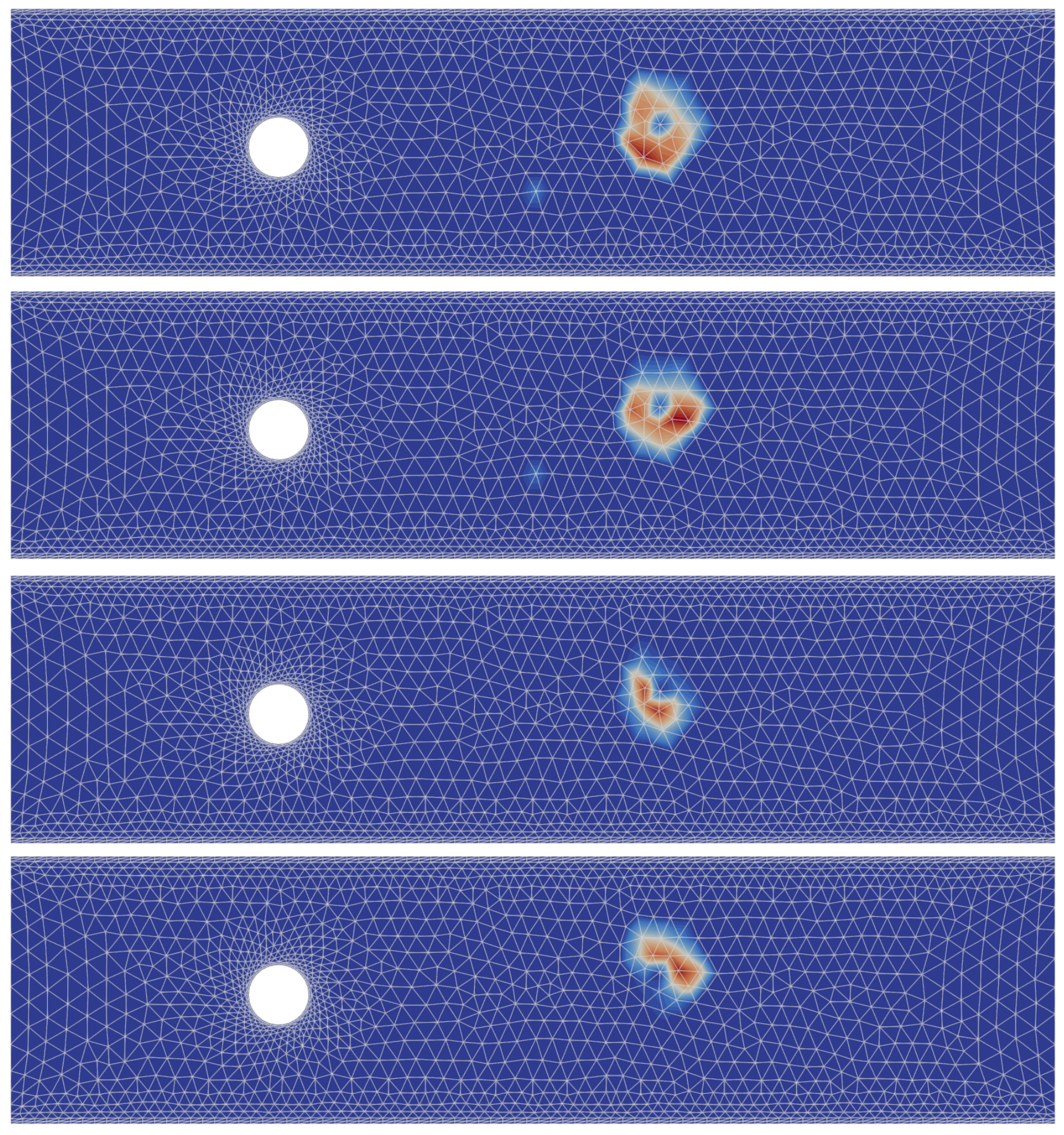}
  \caption{Attention from a single node with the Augmented adjacency matrix. Each row is a different head.}
  \label{fig:attention-cylinder-head}
\end{figure}

We showcase in \autoref{fig:attention-cylinder} the attention from the last transformer block in an \textbf{S} model on the \dataset{Cylinder} dataset. We also demonstrate the difference between three adjacency matrix: $A$, $A^2$ and $A$ with random edges, global attention and 2-Dilation. In \autoref{fig:attention-cylinder-head}, we show the attention with the augmented Adjacency matrix from the 4 different heads. We can indeed notice that each head actually focuses on different part of the receptive field. It is also interesting to notice that the attention magnitude actually largely follows an estimation of the gradient of the velocity.

\section{Predictions}
\label{subappendix:predictions}

\begin{figure}[h!]
  \centering
  \includegraphics[width=0.99\textwidth]{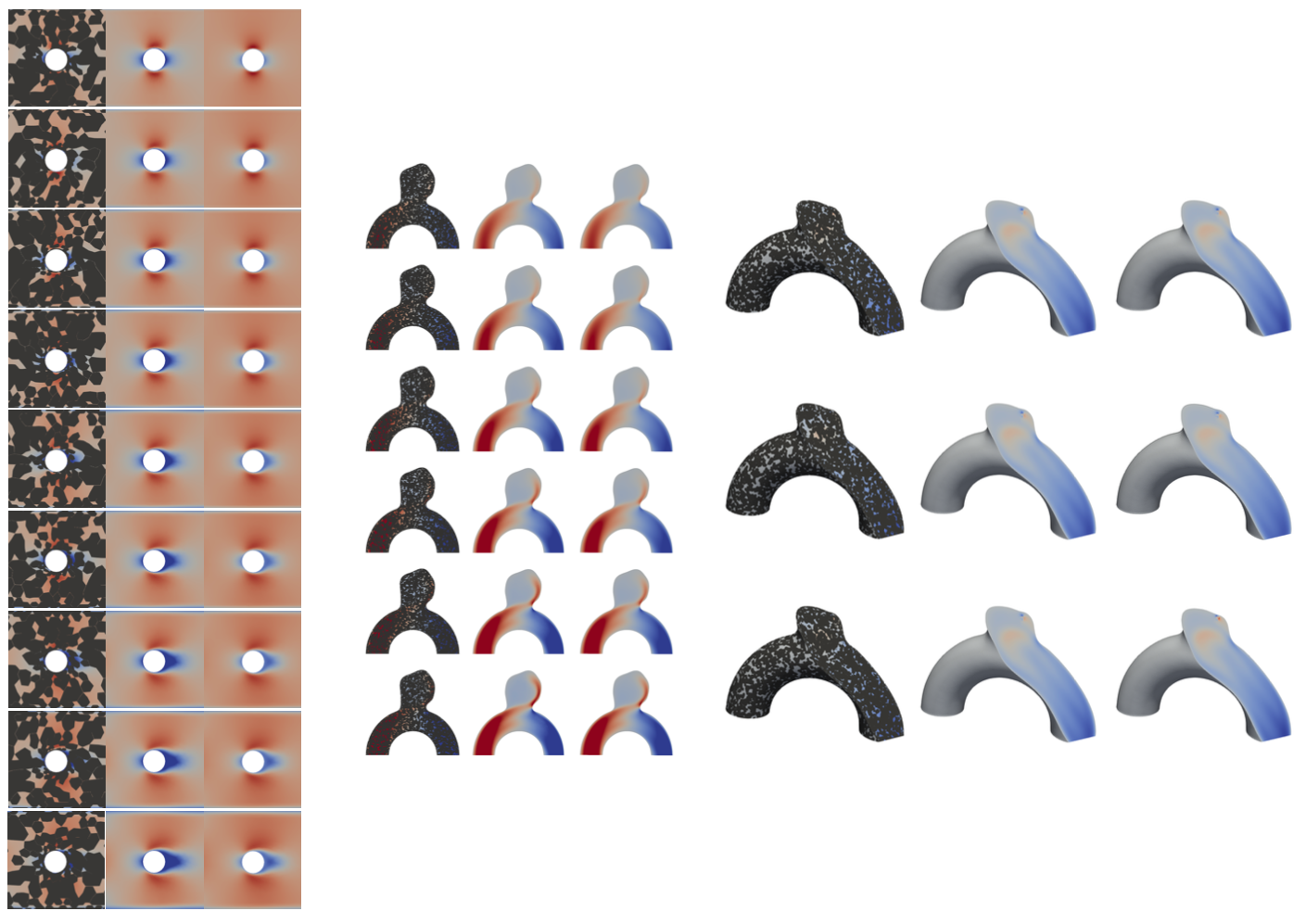}
  \caption{Uncurated random shapes from the validation cylinder, 2D-Aneurysm, and 3D-Aneurysm shapes. \textbf{(left)} masked, \textbf{(middle)} predicted and \textbf{(right)} original.}
  \label{fig:appendix-masking}
\end{figure}

\begin{figure}[h!]
  \centering
  \includegraphics[width=0.99\textwidth]{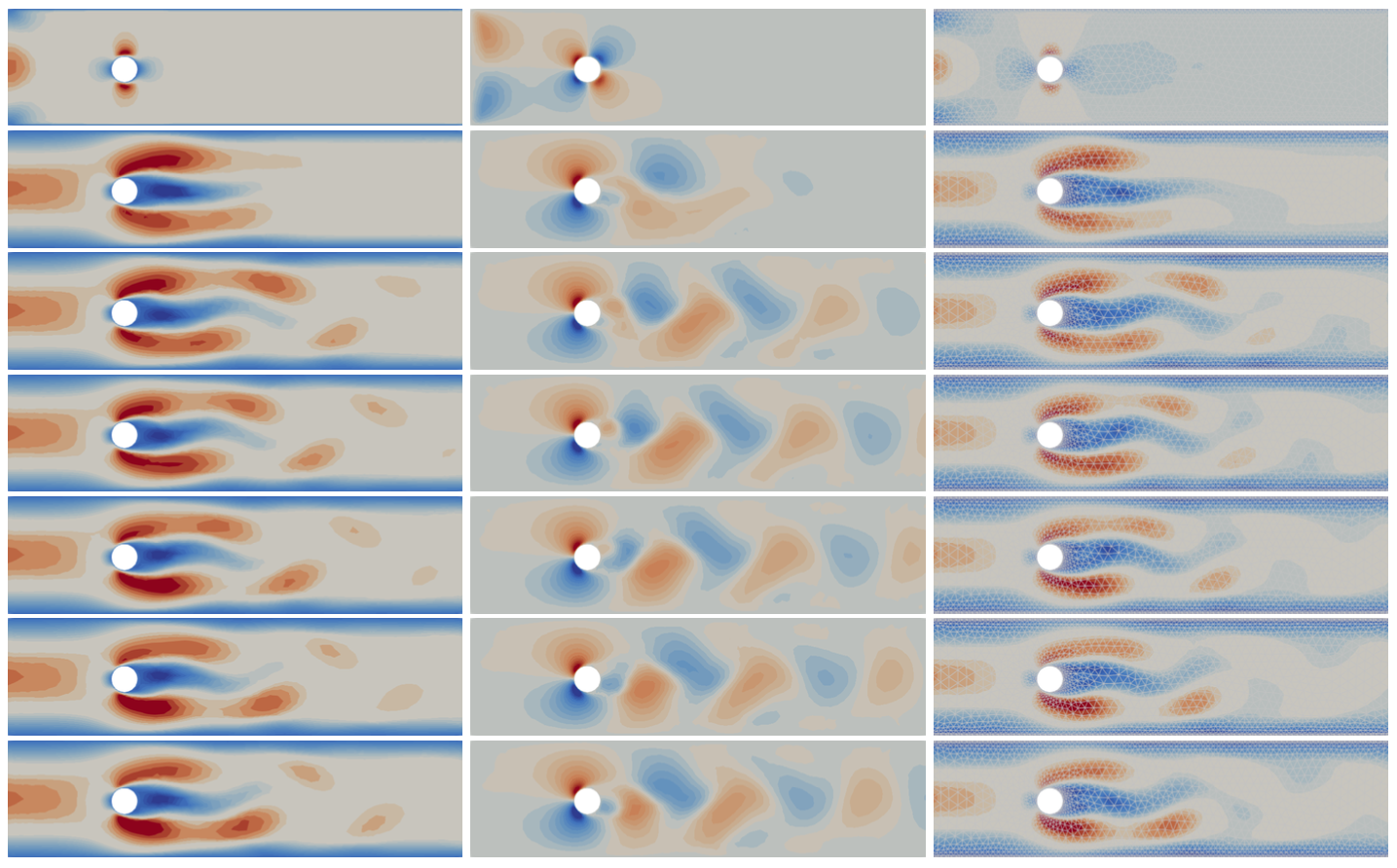}
  \caption{Prediction on one shape from the validation cylinder dataset. \textbf{(left)} $v_x$, \textbf{(middle)} $v_y$ and \textbf{(right)} $||\mathbf{v}||$ displayed on the mesh.}
  \label{fig:appendix-cylinder}
\end{figure}

\begin{figure}[h!]
  \centering
  \includegraphics[width=0.99\textwidth]{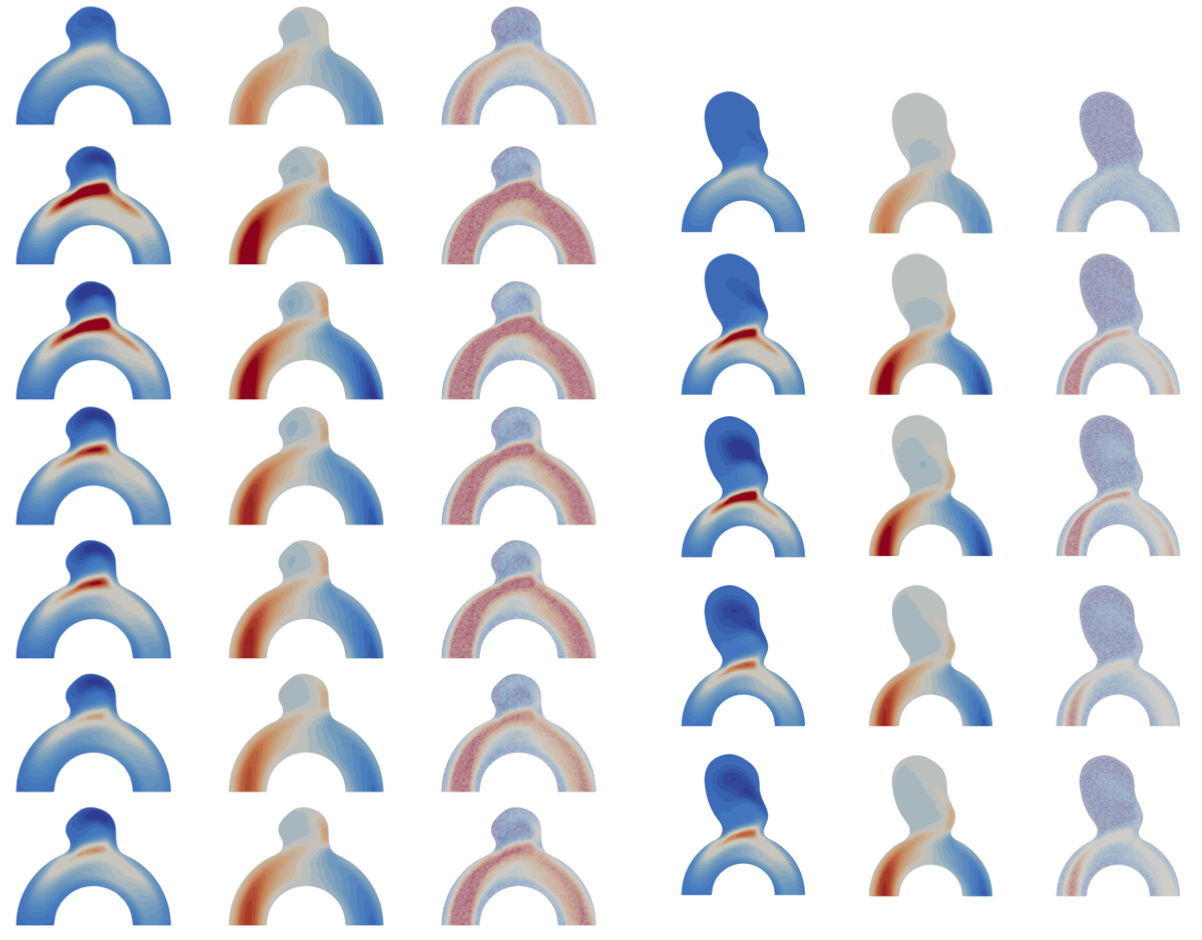}
  \caption{Prediction on 2 shapes from the validation 2D-Aneurysm dataset. \textbf{(left)} $v_x$, \textbf{(middle)} $v_y$ and \textbf{(right)} $||\mathbf{v}||$ displayed on the mesh.}
  \label{fig:appendix-2Daneurysm}
\end{figure}

\begin{figure}[h!]
  \centering
  \includegraphics[width=0.99\textwidth]{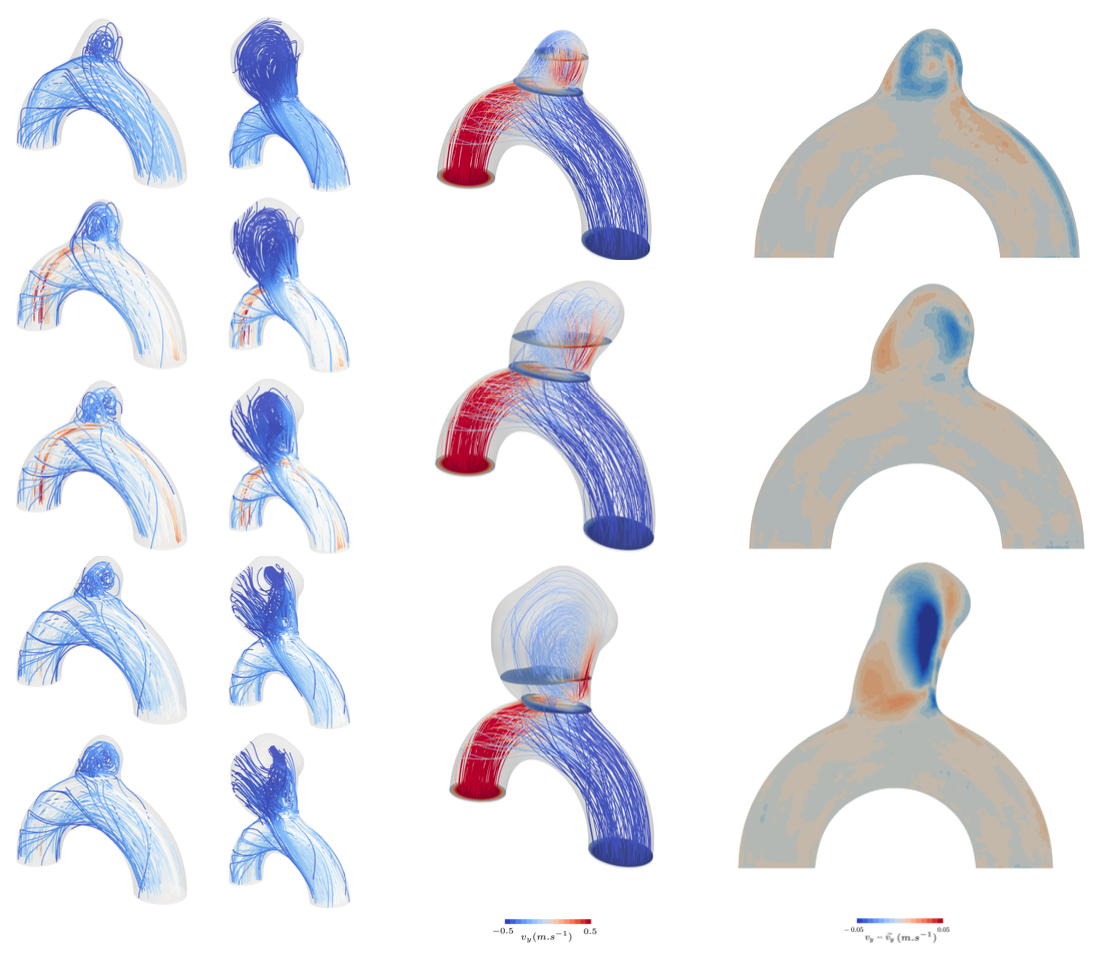}
  \caption{Prediction on 3 shapes from the validation 3D-Aneurysm dataset. \textbf{(left)} $||\mathbf{v}||$, \textbf{(middle)} $v_y$ and \textbf{(right)} the error on $v_y$.}
  \label{fig:appendix-3Daneurysm}
\end{figure}

We display predictions on the \dataset{Cylinder} dataset \autoref{fig:appendix-cylinder}, on the \dataset{2D-Aneurysm} dataset \autoref{fig:appendix-2Daneurysm} and on the \dataset{3D-Aneurysm} dataset \autoref{fig:appendix-3Daneurysm}. Finally, we display the masking pretraining method \autoref{fig:appendix-masking}.

\subsection{Aneurysms}
\label{subappendix:aneurysms}

\begin{figure}[h!]
  \centering
  \includegraphics[width=0.99\textwidth]{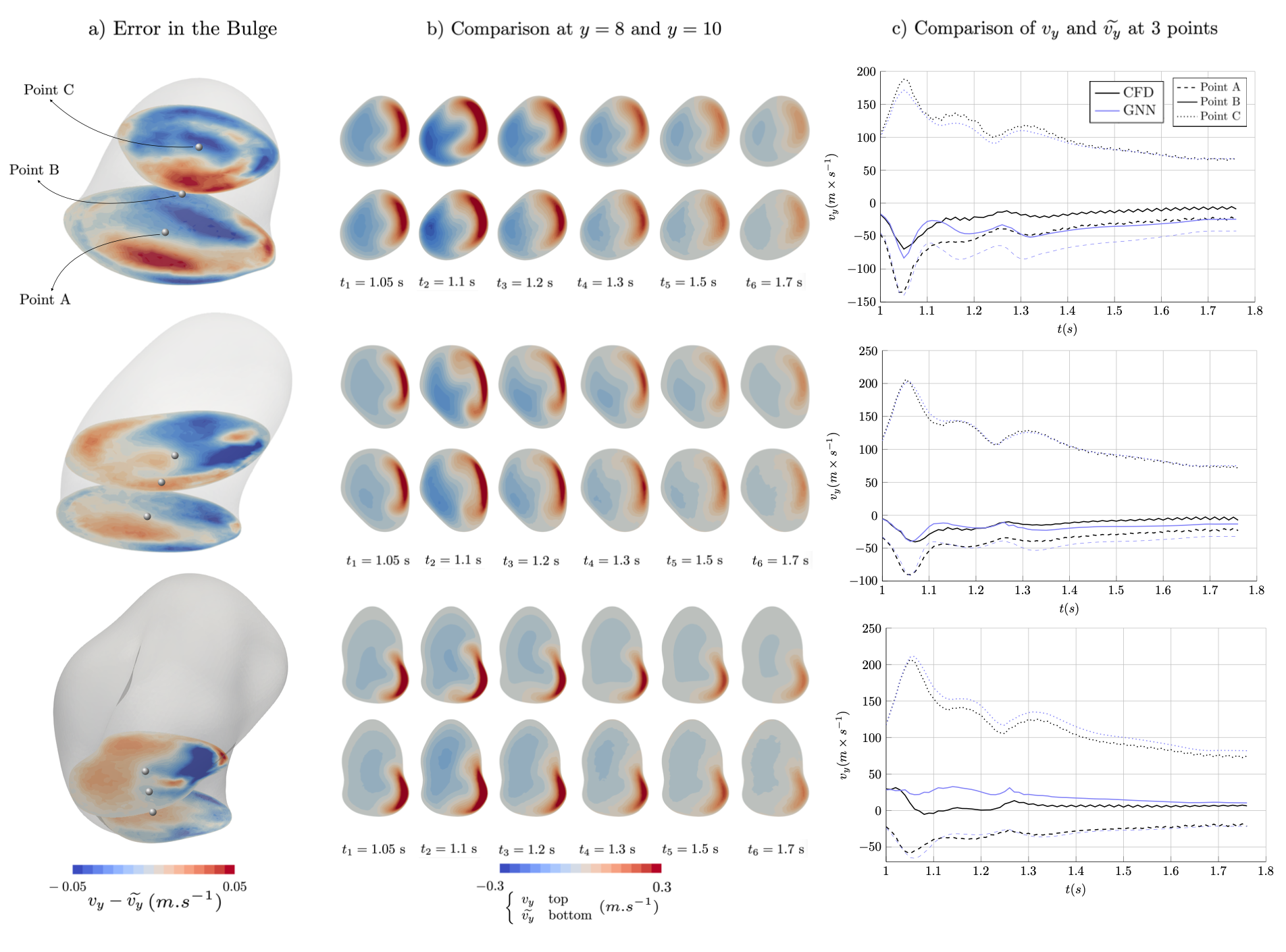}
  \caption{\small\textbf{Flow inside the aneurysm, 2D plan comparison and velocity comparison on three selected points.}
  \textbf{a)} Detailed flow analysis of the selected cases with Systolic flow lines inside the bulge. The two highlighted plans are at $y=8$ and $y=10$. The three selected points used in \textbf{c)} are showcased in gray.
  \textbf{b)} Comparison between the CFD (top row) and our Transformer (bottom row) for $v_y$ inside the bulge in a 2D plan defined by $x=0, y=10, z=0$. We can see the increase and decrease of velocity with the cardiac cycle and a high-fidelity trajectory from our method.
  \textbf{c)} Comparison on 3 selected points at $y=8, y=9, y=10$. Instability in the CFD solution is due to a discrepancy between the timestep used for our training and comparison ($\Delta t = 0.01$) and the one used to solve Navier-Stokes ($\Delta t = 0.002$). Differences in trajectories are within the range of differences between a rigid and an FSI simulation \cite{goetz2023proposal}, showcasing how close to CFD our method is.}
  \label{fig:bulge_compar}
\end{figure}

In \autoref{fig:bulge_compar} we display a detailed comparison between our model and the CFD ground truth on 3 fine aneurysms from the \dataset{3D-Aneurysm }test set. We focus on the velocity field inside the aneurysm since the flow inside the artery is much easier to predict. We select 2 meaningful plans and 3 points inside the aneurysm bulge and compare the 2 velocity fields with a focus on $v_y$. Our model achieves very strong results that are almost impossible to differentiate with a human eye, and show high-fidelity when used for the calculation of clinical metrics such as Wall Shear Stress (WSS).

\section{Non-Physics Datasets}
\label{sec:noncfd}

We also trained our models on four node/graph classification datasets: the MNIST dataset from \cite{dwivedi2022benchmarkinggraphneuralnetworks}, the Reddit dataset from \cite{hamilton2018inductiverepresentationlearninglarge}, the Cora dataset from \cite{McCallum2000} and the PPI dataset from \cite{hamilton2018inductiverepresentationlearninglarge}. 

Results are available in \autoref{table:resultsnoncfd}. For each dataset, we use at least one attention-based architecture and results from the SOTA model for each dataset.

\begin{table*}[!th]
  \centering
  \caption{\label{table:resultsnoncfd}Best results are in \color{red}\textbf{Red}\color{black}, second in \textbf{bold}.}
  \resizebox{0.70\linewidth}{!}
  {
    \begin{tabular}{@{}lllll@{}}
      \toprule
      \sc{Model}                                        & \sc{MNIST} &  \sc{Reddit} &  \sc{Cora} &  \sc{PPI} \\
                                                        & \sc{Accuracy $\uparrow$} & \sc{Accuracy $\uparrow$} & 
                                                        \sc{Accuracy $\uparrow$} &
                                                        \sc{F1 $\uparrow$} \\
                                        \midrule
GAT & — & —   & 84  & 97.3  \\ 
GCN-LPA & — & —  & \textbf{88.5}  & —  \\ 
Graph-Saint & — & \textbf{97}   & —  & \color{red}\textbf{99.5}  \\ 
SAGE & 97.3 & 94.32   & —  & —  \\ 
BNS-GCN & — & \color{red}\textbf{97.17}   & —  & —  \\
NeuralWalker & \color{red}\textbf{98.76} & —   & —  & —  \\
\midrule
\textbf{Ours} & \textbf{98.2} & \textbf{97}  & \color{red}\textbf{88.6}   & \textbf{99.2}  \\            
      \bottomrule
    \end{tabular}
  }
\end{table*}

\end{document}